%% file: main.tex
\documentclass[10pt,twocolumn,letterpaper]{article}

\usepackage{cvpr}              

\definecolor{smoothred}{RGB}{220, 50, 47} 
\definecolor{smoothblue}{RGB}{68, 90, 225} 
\definecolor{smoothgreen}{RGB}{80, 160, 90}
\definecolor{VisualPerception}{RGB}{110,155,170}   
\definecolor{TextualReasoning}{RGB}{140,125,190}   
\definecolor{text-green}{RGB}{91,124,131}
\definecolor{text-yellow}{RGB}{168,146,89}

\definecolor{table-yellow}{RGB}{250, 223, 161}
\definecolor{table-blue}{RGB}{173, 216, 230}
\definecolor{table-green}{RGB}{126, 172, 181}
\definecolor{darkblue}{rgb}{0, 0, 0.5}
\definecolor{PDCRColor}{rgb}{0.0, 0.5, 0.5}

\usepackage{wrapfig}
\usepackage{comment}
\usepackage{nicefrac}       
\usepackage{microtype}      
\usepackage{xcolor}         
\usepackage{colortbl} 
\usepackage{subcaption}
\usepackage{multirow}
\usepackage{verbatim}
\usepackage{diagbox}
\usepackage{bm}
\usepackage{makecell}
\usepackage{graphicx}
\usepackage{amsmath}
\usepackage[english]{babel}
\usepackage{mathtools}
\usepackage[ruled,vlined]{algorithm2e}
\usepackage[utf8]{inputenc}
\usepackage{csquotes}
\usepackage{enumitem}
\usepackage{soul}
\usepackage{lipsum} 

\usepackage{booktabs} 
\usepackage{caption} 
\usepackage{rotating} 
\usepackage{multirow} 
\usepackage{lipsum} 
\usepackage{colortbl}
\usepackage{amsthm}
\usepackage{amssymb}
\usepackage{bbm}

\usepackage[most]{tcolorbox}
\tcbuselibrary{breakable, skins}
\usepackage{longtable}
\usepackage{booktabs}

\usepackage{arydshln}

\input{preamble}

\definecolor{cvprblue}{rgb}{0.21,0.49,0.74}
\usepackage[pagebackref,breaklinks,colorlinks,allcolors=cvprblue]{hyperref}

\def\paperID{38938} 
\def\confName{CVPR}
\def\confYear{2026}

\title{PDCR: Perception-Decomposed Confidence Reward for Vision-Language Reasoning}

\author{%
  Hee Suk Yoon${}^1$\thanks{Equal Contribution.} \thanks{This work was done while Hee Suk Yoon was an intern at MSRA.}\,\,\,\,\,\,Eunseop Yoon${}^1$${}^*$\,\,\,\,Ji Woo Hong${}^1$\,\,\,\,SooHwan Eom${}^1$\,\,\,\,Gwanhyeong Koo${}^1$\\
  Mark Hasegawa-Johnson${}^2$\,\,\,\,Qi Dai${}^3$\,\,\,\,Chong Luo${}^3$${}^\text{\ddag}$\,\,\,\,Chang D. Yoo${}^1$\thanks{Corresponding Authors.} \\
  ${}^1$Korea Advanced Institute of Science and Technology (KAIST)\\${}^2$University of Illinois at Urbana-Champaign (UIUC)\,\,\,\,\,\,\,\,${}^3$Microsoft Research Asia (MSRA)\\
  \small{\texttt{\{hskyoon,esyoon97,cd\_yoo\}@kaist.ac.kr},\,\,\,\,\,\,\,\texttt{\{jhasegaw\}@illinois.edu},\,\,\,\,\,\,\,\texttt{\{qid,clou\}@microsoft.com}} \\
}

\begin{document}
\maketitle
\input{sec/0_abstract}    
\input{sec/1_intro}
\input{sec/2_related_works}
\input{sec/3_background}
\input{sec/4_observations}
\input{sec/5_method}
\input{sec/6_experiments}
\input{sec/7_conclusion}
\clearpage
\section{Acknowledgment}
This research was supported by the MSIT(Ministry of Science, ICT), Korea, under the Global Research Support Program in the Digital Field program(RS-2024-00436680) supervised by the IITP(Institute for Information \& Communications Technology Planning \& Evaluation). This project is supported by Microsoft Research Asia.
{
    \small
    \bibliographystyle{ieeenat_fullname}
    \bibliography{main}
}

\onecolumn
\input{sec/X_suppl}

\end{document}

%% file: preamble.tex
\newcommand{\red}[1]{{\color{red}#1}}

\usepackage{titletoc}

\titlecontents{section}
  [0em]                             
  {\addvspace{1.0\baselineskip}}    
  {\makebox[2em][l]{\thecontentslabel}} 
  {}                                
  {\hfill\contentspage}             

\titlecontents{subsection}
  [1.5em]
  {\addvspace{0.2\baselineskip}}
  {\makebox[3em][l]{\thecontentslabel}} 
  {}
  {\hfill\contentspage}             

%% file: sec/0_abstract.tex
\begin{abstract}
Reinforcement Learning with Verifiable Rewards (RLVR) traditionally relies on a sparse, outcome-based signal. Recent work shows that providing a fine-grained, model-intrinsic signal—rewarding the confidence growth in the ground-truth answer—effectively improves language reasoning training by providing step-level guidance without costly external models. While effective for unimodal text, we find that naively applying this global reward to vision-language (V-L) reasoning is a suboptimal strategy, as the task is a heterogeneous mix of sparse visual perception and dense textual reasoning. This global normalization creates mixture-induced signal degradation, where the training signal for visual steps is statistically distorted by the predominant textual steps.
We propose \textbf{Perception-Decomposed Confidence Reward (PDCR)}, a framework that solves this by aligning the reward structure with the task's heterogeneous nature. PDCR first performs an unsupervised skill decomposition, introducing a model-internal Visual Dependence Score to quantify visual reliance and applying a clustering algorithm to separate perception and reasoning steps. Based on this, PDCR computes a decomposed advantage by normalizing confidence gains within each skill cluster. This intra-cluster normalization provides a stable, correctly-scaled signal for both perception and reasoning. We demonstrate that PDCR outperforms the naive, global-reward formulation and sparse-reward baselines on key V-L reasoning benchmarks. The code is publicly accessible at \href{https://github.com/hee-suk-yoon/PDCR}{https://github.com/hee-suk-yoon/PDCR}.
\end{abstract}

%% file: sec/1_intro.tex
\section{Introduction}
\label{sec:intro}
\begin{figure}[t]
  \centering
   \includegraphics[width=1.0\linewidth]{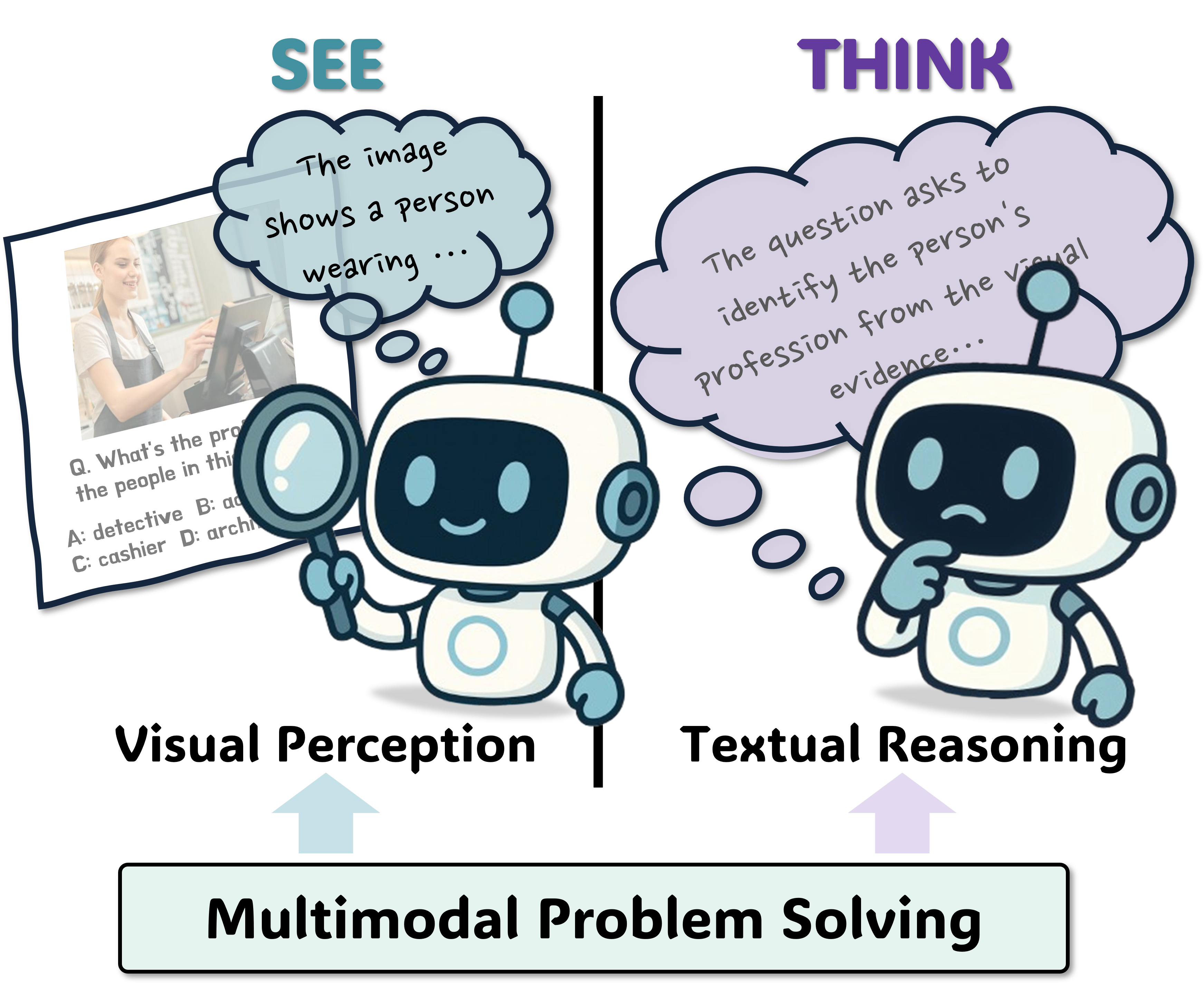}
\caption{Multimodal reasoning mixes two distinct behaviors: seeing (visual perception, extracting evidence from the image) and thinking (textual reasoning over that evidence). Our work argues that these heterogeneous skills must be rewarded independently, as a naive, global reward signal fails to properly assign credit to each.} 
   \label{fig:intro}
   \vspace{-0.5cm}
\end{figure}
Pre-trained vision-language models (VLMs) exhibit strong performance on complex, multi-step reasoning tasks \cite{visual_instruction_tuning, team2023gemini, 2023GPT4VisionSC, bai2025qwen2, yang2025qwen3}. Reinforcement Learning with Verifiable Rewards (RLVR) has emerged as a leading approach for further improving these capabilities, using a programmatically checkable terminal metric (e.g., correctness of the final answer) as the reward \cite{qwen2_5_math, deepseekr1, grpo, huang2025vision, xia2025visionary}. While effective, the standard RLVR formulation supplies a sparse terminal accuracy signal, offering no guidance for intermediate steps and thus creating a credit assignment problem.

To address this, recent work has explored fine-grained, step-level supervision. One prominent approach is training an external Process Reward Model (PRM) \cite{yoon2024tlcr, prm1, prm2, prm3, prm_mllm}. However, PRMs are often costly to train, data-hungry, and prone to misalignment. In contrast, a compelling, model-intrinsic alternative from the language domain is to reward the ground-truth confidence growth \cite{yoon2025pacr}. This method provides step-level guidance using the model's own probabilities of the ground-truth answer, avoiding the need for an external reward model to evaluate each reasoning step.

While effective for unimodal text, we find that naively applying this global reward to \textbf{vision-language (V-L) reasoning} is a suboptimal strategy. As illustrated in Figure \ref{fig:intro}, the core challenge is that V-L reasoning is not a homogeneous process; it is a \textbf{heterogeneous mixture of two distinct, high-level skills}: (i) sparse, but critical, \textcolor{VisualPerception}{\textbf{visual perception}} steps and (ii) dense, predominant \textcolor{TextualReasoning}{\textbf{textual reasoning}} steps. This statistical imbalance, as we show in Section~\ref{sec:challenge}, leads directly to mixture-induced signal degradation: the reward signal for critical visual steps is statistically distorted by the predominant textual steps.

To this end, we propose Perception-Decomposed Confidence Reward (PDCR), a framework that solves this by aligning the reward structure with the task's heterogeneous nature. As illustrated in Figure~\ref{fig:pdcr_method}, PDCR operates in two stages:
\begin{enumerate}
\item \textbf{Unsupervised Skill Decomposition:} We first introduce a Visual Dependence Score, a model-internal signal that quantifies a step's reliance on the visual input. We then apply an optimal clustering algorithm (Otsu's method) to automatically separate steps into \textcolor{VisualPerception}{\textbf{visual perception}} and \textcolor{TextualReasoning}{\textbf{textual reasoning}} groups.
\item \textbf{Perception-Decomposed Advantage:} Based on this, PDCR computes a decomposed advantage by normalizing confidence gains \textit{within} each respective skill cluster, ensuring a stable, intra-cluster comparison.
\end{enumerate}

This decomposed approach, when paired with the standard terminal reward, provides a more robust and correctly-scaled training signal. By calculating a separate advantage for visual perception and textual reasoning, PDCR ensures that each skill is evaluated against its own statistical baseline, allowing visual perception steps to receive a stable, well-scaled signal that is not distorted by the predominant textual reasoning steps. 

\textbf{In detail, our contributions can be summarized as follows:}
\begin{itemize}
\item Through a series of observations, we identify the \textbf{mixture-induced signal degradation} problem in vision-language reasoning, showing that naive, model-intrinsic rewards are \textbf{suboptimal} due to skill heterogeneity.
\item We propose an \textbf{unsupervised skill decomposition} method using a \textbf{Visual Dependence Score} and optimal clustering, which requires no external labels during training.
\item We introduce \textbf{PDCR}, a decomposed reward framework that uses this clustering to compute stable, intra-cluster advantages, resolving the signal degradation problem.
\item We demonstrate through experiments on key vision-language reasoning benchmarks that our PDCR framework outperforms both sparse-reward baselines and the naive, global-reward formulation.
\end{itemize}

%% file: sec/2_related_works.tex
\section{Related Works}
\label{sec:related_works}

\subsection{Reasoning in Multimodal Models}
Improving the reasoning capabilities of Multimodal Large Language Models (MLLMs) is an active area of research \cite{kim2021structured, yoon2023hear, huang2025vision, xia2025visionary, huang2025spotlight, fan2025sophiavl, liu2025visionreasoner, sarch2025grounded, xu2025defacto, qiu2025noisygrpo, ong2025training, chu2025qwen, yoon2024bi}. A common paradigm is to fine-tune MLLMs using Reinforcement Learning (RL) with verifiable rewards \cite{deepseekr1, grpo, qwen2_5_math}. However, this approach typically relies on a sparse, outcome-based reward (i.e., whether the final answer is correct), which is inefficient for training complex, multi-step tasks.

To overcome this, several works have introduced external Process Reward Models (PRMs) \cite{prm1, prm2, prm3, prm_mllm}. PRMs provide a dense, step-level training signal, but are notoriously costly to train, often requiring extensive and difficult dataset curation. This cost has motivated a recent shift towards finding reward signals that are model-intrinsic or self-generated.

\subsection{Model-Intrinsic and Self-Rewarding Methods}
To address the costs of external PRMs, an alternative direction is self-rewarding reinforcement learning, which sources rewards from the model's own internal signals \cite{yuan2024self, zhou2024calibrated, yoon2025confpo}. These signals often include logit-based self-certainty \cite{zhao2025learning}, self-confidence \cite{li2025confidence}, or majority-voting \cite{zhou2025evolving}.

A particularly relevant framework in the language domain is the Progressively Ascending Confidence Reward (PACR) \cite{yoon2025pacr}. PACR provides a dense, step-level reward by calculating the increase in the model's log-probability of the ground-truth answer, guided by the inductive bias that confidence should ascend during a correct reasoning trace. This approach has proven effective for text-only reasoning but has not been sufficiently adapted for the unique challenges of multimodal tasks.

\subsection{Decomposed Reward Modeling}
A parallel line of research focuses on decomposing complex reward signals. In text-only LLMs, this is often done by evaluating a response from various perspectives, such as helpfulness, safety, or correctness \cite{finegrained_rlhf, ultrafeedback, openprm}. This concept is also being explored in MLLMs, where the reward signal is often denser due to the multimodal inputs \cite{xia2025visionary, ozdemir2024enhancing}. 

However, existing work on vision-language decomposition primarily focuses on generating explicit textual descriptions (e.g., image captions) to serve as the basis for the reward \cite{sr1, liu2025visionreasoner, ding2025vtperception}. Our work diverges from this approach by decomposing the original reasoning steps directly via an internal, process-level signal, rather than relying on the generation of additional, explicit text like an image caption. 

We argue that visual and linguistic capabilities are heterogeneous skills with distinct distributions, and therefore benefit from independent normalization. Analogous concepts have been explored in concurrent works in other domains, such as multi-reward \cite{liu2026gdpo} or tool-use agentic tasks \cite{zhu2025stratified}.
\begin{figure}[t!]
  \centering
   \includegraphics[width=0.95\linewidth]{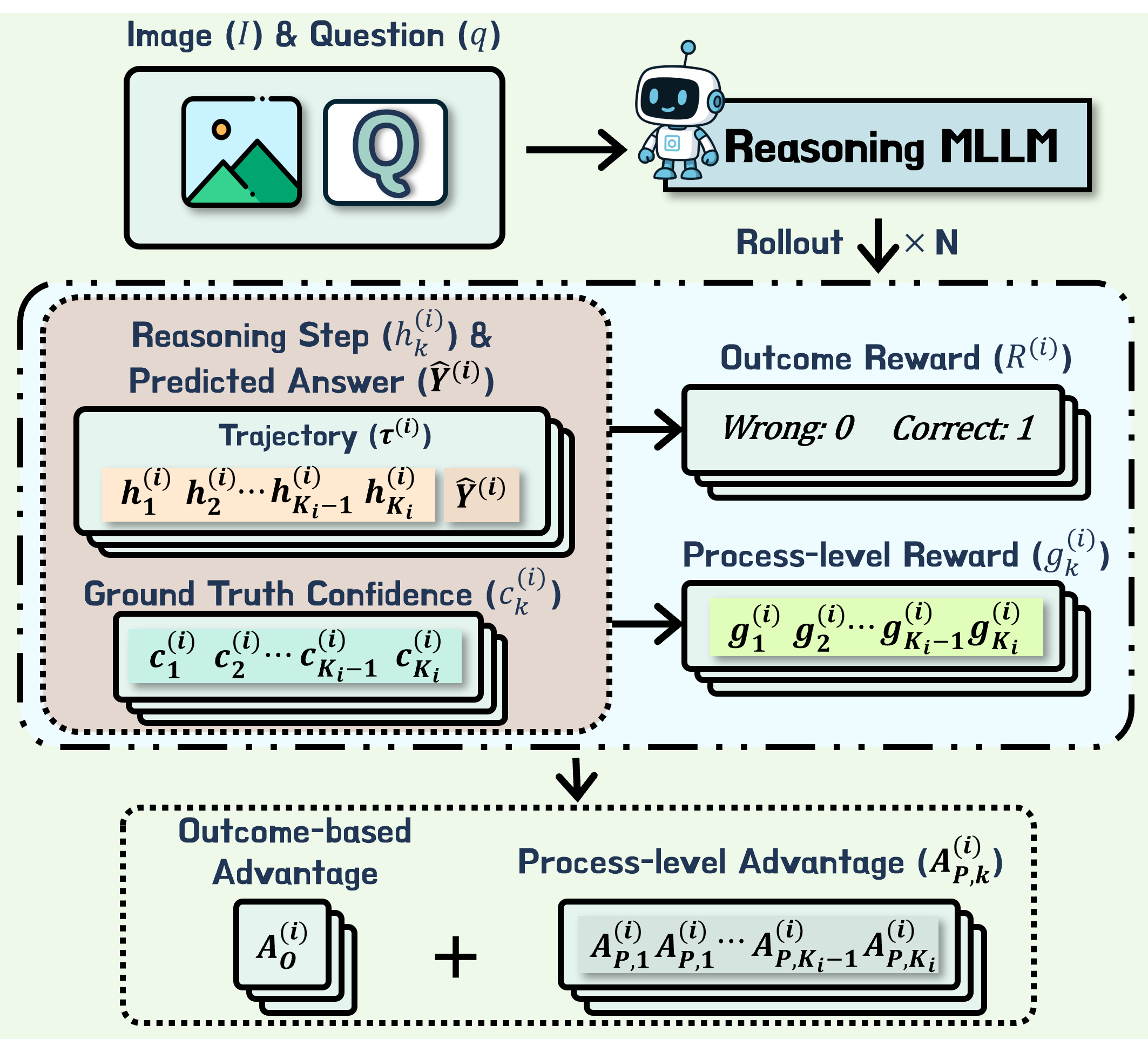}
   \caption{The baseline dense reward pipeline. For $N$ rollouts, a sparse \textbf{Outcome Reward} ($R^{(i)}$) is computed from the final answer's correctness. Concurrently, the model's stepwise ground-truth confidence ($c_k^{(i)}$) is used to derive a dense \textbf{Process-level Reward} ($g_k^{(i)}$) (i.e., the confidence gain). These two rewards are then converted into an Outcome-based Advantage ($A_O^{(i)}$) and a globally-normalized Process-level Advantage ($A_{P,k}^{(i)}$).} 
   \label{fig:background}
\end{figure}

%% file: sec/3_background.tex
\section{Background and Problem Setup}
This section first describes the standard sparse-reward RL framework (GRPO) for reasoning training. It then discusses its sparse-reward limitations, followed by the recent, model-intrinsic signal of ground-truth confidence growth \cite{yoon2025pacr} which addresses them.
 
\subsection{Group Relative Policy Optimization (GRPO)} As illustrated in Figure~\ref{fig:background}, for a given visual input $\mathbf{I}$ and question $\mathbf{q}$, the behavior policy $\pi_{\theta_{\text{old}}}$ generates a group of $N$ reasoning trajectories $\{\tau^{(i)}\}_{i=1}^N$. Each trajectory $\tau^{(i)}$ consists of multiple reasoning steps $\{h^{(i)}_k\}_{k=1}^{K_i}$, and the final predicted answer $\hat{Y}^{(i)}$.

For each trajectory $\tau^{(i)}$, a sparse, terminal outcome reward $R^{(i)}$ is assigned based on the final answer's correctness:
\begin{equation}
    \label{eq:sparse_reward}
    R^{(i)} = \begin{cases} 1, & \text{if } is\_equivalent(\hat{Y}^{(i)}, Y_{gt}) \\ 0, & \text{otherwise.} \end{cases}
\end{equation}
This sparse reward is then used to compute the outcome-based advantage $A_{O}^{(i)}$ by normalizing it with the group's mean and standard deviation:
\begin{equation}
    \label{eq:sparse_advantage}
    A_{O}^{(i)} = \frac{R^{(i)} - \text{mean}(\{R^{(j)}\}_{j=1}^{N})}{\text{std}(\{R^{(j)}\}_{j=1}^{N})}.
\end{equation}
This single, trajectory-level advantage $A_{O}^{(i)}$ is then applied to all tokens within all reasoning steps $h_k^{(i)}$ of the trajectory $\tau^{(i)}$. This sparse signal offers no step-specific guidance, thus exacerbating the credit assignment problem.
\subsection{Ground-Truth Confidence Growth as a Model-Intrinsic Process Reward}
\label{sec:pacr}
To address the limitations of this sparse, trajectory-level signal, recent work in the language domain \cite{yoon2025pacr} introduced a dense, model-intrinsic reward based on \textbf{ground-truth confidence growth}. This approach is guided by the inductive bias that a model's confidence in the correct answer $Y_{gt}$ should progressively increase as it generates a well-formed reasoning trajectory.

To operationalize this inductive bias as a computable reward, each trajectory $\tau^{(i)}$ is first segmented into its discrete reasoning steps $\{h^{(i)}_k\}_{k=1}^{K_i}$ (see Appendix~\ref{app:segmentation} for segmentation details). As shown in the overall pipeline in Figure~\ref{fig:background}, this segmentation allows for a process-level reward. At each step $k$ of trajectory $\tau^{(i)}$, the model's confidence in the ground truth $c^{(i)}_k$ is evaluated as its log-probability given the current prefix:
\begin{equation}
    \label{eq:step_confidence}
    c^{(i)}_k = \log \pi_{\theta}(Y_{gt} | \mathbf{I}, \mathbf{q}, H^{(i)}_{\le k}),
\end{equation}
where $H^{(i)}_{\le k} = (h^{(i)}_1, \dots, h^{(i)}_k)$. The core reward signal is the \textit{stepwise confidence gain} $g^{(i)}_k$, defined as the change in confidence induced by the addition of step $h_k$:
\begin{equation}
    \label{eq:confidence_gain}
    g^{(i)}_k := c^{(i)}_k - c^{(i)}_{k-1},
\end{equation}
where $c^{(i)}_0 = \log \pi_{\theta}(Y_{gt} | \mathbf{I}, \mathbf{q})$. A positive gain $g^{(i)}_k > 0$ indicates that step $h_k$ made the ground truth more probable.

\begin{figure*}[t!]
  \centering
   \includegraphics[width=1.0\linewidth]{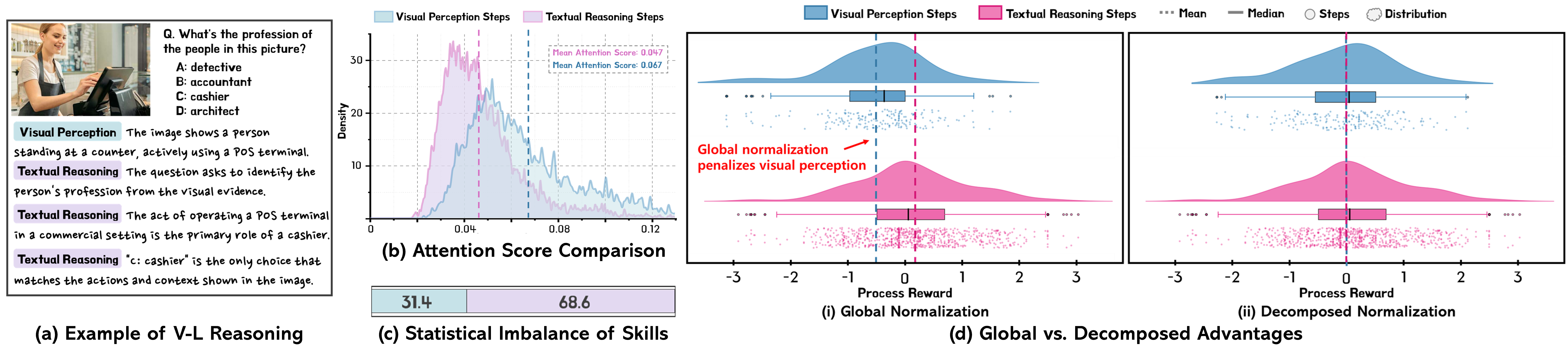}
\caption{An illustration of our core observations and the mixture-induced signal degradation problem.
\textbf{(a)} A V-L reasoning trace is a heterogeneous mix of \textcolor{VisualPerception}{visual perception} steps and \textcolor{TextualReasoning}{textual reasoning} steps.
\textbf{(b)} We validate this functional distinction by analyzing attention: \textcolor{VisualPerception}{perception} steps attend highly to visual tokens, while \textcolor{TextualReasoning}{reasoning} steps do not.
\textbf{(c)} These skills are statistically imbalanced: \textcolor{VisualPerception}{perception} steps are sparse (31.4\%), while \textcolor{TextualReasoning}{reasoning} steps form the dense majority (68.6\%).
\textbf{(d)} This imbalance causes \textit{mixture-induced signal degradation}: (i) a naive, global normalization (Eq.~\ref{eq:pacr_advantage}) compresses and misaligns the advantage distribution for perception steps. (ii) Our proposed decomposed normalization provides a stable, well-scaled advantage for both skills.}
   \label{fig:observation}
\end{figure*}

This stepwise gain $g^{(i)}_k$ is used to compute a dense, process-level advantage. In PACR \cite{yoon2025pacr}, the gains are first converted into a discounted return $G_{k}^{(i)} = \sum_{m=k}^{K_i} \gamma^{m-k} g_{m}^{(i)}$. This discounted return is then normalized against the statistics (e.g., min/max) of a global pool comprising the returns of all steps from all $N$ trajectories. Let $I = \{(j, k') \mid 1 \le j \le N, 1 \le k' \le K_j\}$ be the set of all step indices in the group. The process-level advantage $A_{P,k}^{(i)}$ is:
\begin{equation}
    \label{eq:pacr_advantage}
    A_{P,k}^{(i)} = \frac{G_{k}^{(i)} - \min\limits_{(j,k') \in I} G_{k'}^{(j)}}{\max\limits_{(j,k') \in I} G_{k'}^{(j)} - \min\limits_{(j,k') \in I} G_{k'}^{(j)}}.
\end{equation}
Finally, this process-level advantage is combined with the sparse, outcome-based advantage from Eq.~\ref{eq:sparse_advantage}:
\begin{equation}
    \label{eq:combined_advantage}
    A_{k,total}^{(i)} = \lambda_{O} A_{O}^{(i)} + \lambda_{P} A_{P,k}^{(i)},
\end{equation}
where $\lambda_{O}$ and $\lambda_{P}$ are weighting hyperparameters. During the policy update, this final step-level advantage $A_{k,total}^{(i)}$ is then applied to all tokens that comprise the reasoning step $h_k^{(i)}$. This combined signal provides fine-grained, step-specific guidance, effectively solving the credit assignment problem of sparse-only rewards in the language domain, while avoiding the need for external reward models.

%% file: sec/4_observations.tex
\section{The Challenge of Heterogeneous Skills in Vision-Language Reasoning}
\label{sec:challenge}

While the dense reward signal from ground-truth confidence growth (Eq.~\ref{eq:pacr_advantage}) is effective for unimodal language tasks, its naive application to vision-language (V-L) reasoning is suboptimal. The core issue, which we validate in this section, is that V-L reasoning is not a \textit{homogeneous} process. Instead, it is a \textit{heterogeneous} mixture of two functionally distinct, high-level skills: (1) sparse, but critical, \textcolor{VisualPerception}{\textbf{visual perception}} steps and (2) dense, predominant \textcolor{TextualReasoning}{\textbf{textual reasoning}} steps.

\paragraph{\texttt{[Observation 1]}: V-L Reasoning as a Heterogeneous Skill Mix.}

The two primary skills in a V-L reasoning trajectory serve different purposes. As illustrated in the example in Figure~\ref{fig:observation}-(a), \textcolor{VisualPerception}{\textbf{visual perception}} involves extracting evidence from the visual input $\mathbf{I}$ and translating it into textual observations. This includes identifying objects, their attributes, or their spatial relationships (e.g., \textit{``The image shows a person standing at a counter, actively using a POS terminal."}). In contrast, \textcolor{TextualReasoning}{\textbf{textual reasoning}} involves logical deduction, calculation, or conclusion-drawing based on previously established facts (e.g., \textit{``The act of operating a POS terminal in a commercial setting is the primary role of a cashier."}).

To empirically validate this functional distinction, we analyze the model's internal attention patterns. We first annotated a representative set of reasoning trajectories, categorizing each step as either \textcolor{VisualPerception}{\textbf{visual perception}} or \textcolor{TextualReasoning}{\textbf{textual reasoning}} (see Appendix~\ref{app:gpt_label} for full annotation details). We then measured the average attention each step paid to the \textit{input visual tokens}. As shown in the plot in Figure~\ref{fig:observation}-(b), the two skills exhibit distinct attention profiles. \textcolor{VisualPerception}{\textbf{Visual perception}} steps exhibit high attention to the input visual tokens, as they are responsible for converting visual information into a textual format. Conversely, \textcolor{TextualReasoning}{\textbf{textual reasoning}} steps attend almost exclusively to the preceding text tokens, using the information established by prior steps to perform logical inference. This shows that a V-L reasoning trajectory is not a homogeneous process, but a heterogeneous mixture of these two distinct capabilities.

\paragraph{\texttt{[Observation 2]}: Statistical Imbalance of Skills.}
Beyond their functional differences, we find a statistical imbalance between the two skills. Analyzing the same ground-truth annotated set from \texttt{[Observation 1]}, we observe that \textcolor{VisualPerception}{\textbf{visual perception}} steps are inherently sparse, accounting for approximately 30\% of the total reasoning steps, while the \textcolor{TextualReasoning}{\textbf{textual reasoning}} steps form the dense majority (approx. 70\%), as shown in Figure~\ref{fig:observation}-(c). This sparsity is a key driver of the normalization problem we describe next.

\paragraph{\texttt{[Observation 3]}: Mixture-Induced Signal Degradation from Skill Imbalance.}
This statistical imbalance poses a significant challenge for the global normalization in Eq.~\ref{eq:pacr_advantage}. We term this issue mixture-induced signal degradation, as the global statistics ($\min, \max$) are computed from a pool overwhelmingly dominated by the returns of \textcolor{TextualReasoning}{\textbf{textual}} steps. This global baseline is, therefore, unrepresentative of the sparse \textcolor{VisualPerception}{\textbf{perception}} steps.

\begin{figure*}[t!]
  \centering
   \includegraphics[width=1.0\linewidth]{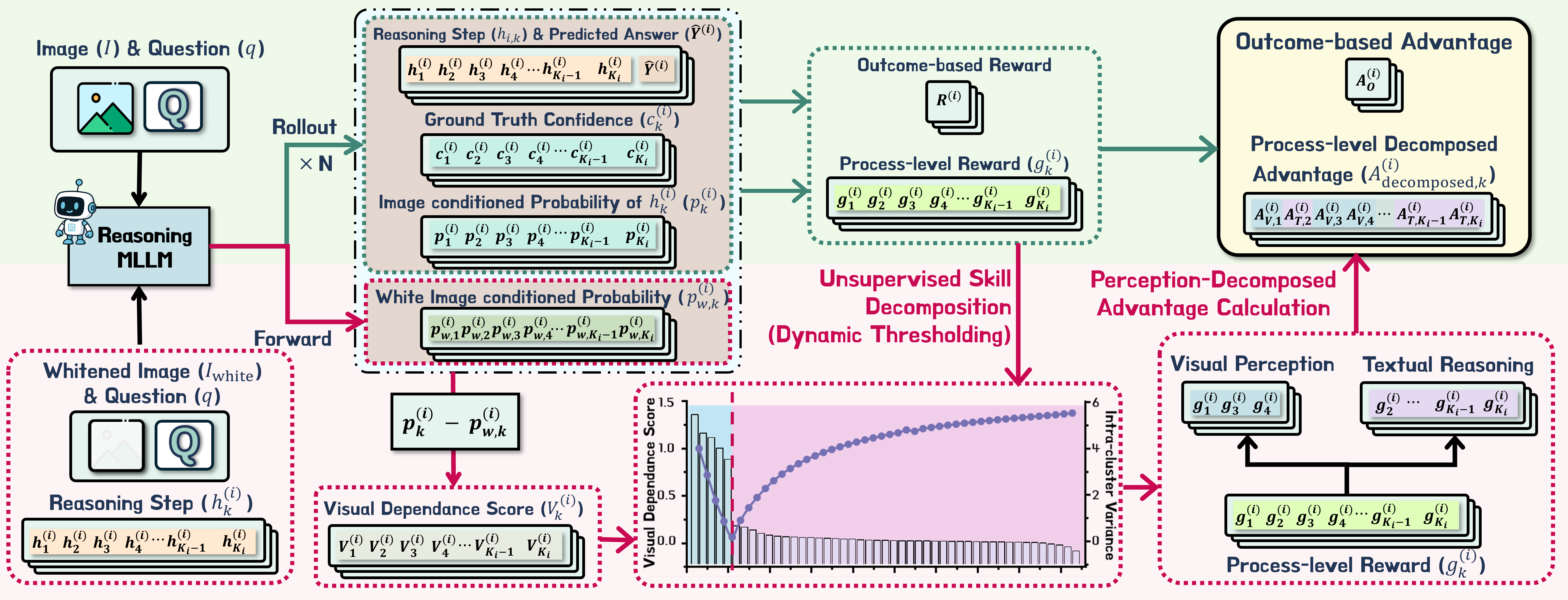}
\caption{Overview of our PDCR (Perception-Decomposed Confidence Reward) framework. \textbf{(Top, green path)}: We compute the standard Process-level Reward ($g_k^{(i)}$) (i.e., the confidence gain, Eq.~\ref{eq:confidence_gain}) from the ground-truth confidence $c_k^{(i)}$. \textbf{(Bottom, pink path)}: We perform \textbf{Unsupervised Skill Decomposition}. We introduce a \textbf{Visual Dependence Score ($V_k^{(i)}$)} (Eq.~\ref{eq:vis_dependence}) by comparing the log-probability of a step given the real image ($\mathbf{I}$) versus a blank image ($\mathbf{I}_{\text{white}}$). We apply an optimal clustering algorithm (Otsu's method, Eq.~\ref{eq:argmin_sse}) to these scores to separate all steps $h_k^{(i)}$ into \textit{Visual Perception} ($\mathcal{I}_{\text{visual}}$) and \textit{Textual Reasoning} ($\mathcal{I}_{\text{textual}}$) groups. \textbf{(Right)}: In the final \textbf{Perception-Decomposed Advantage Calculation}, the reward $g_k^{(i)}$ from the top path is normalized \textit{within} the skill cluster identified by the bottom path (Eq.~\ref{eq:vis_advantage} \& \ref{eq:tex_advantage}). This yields the decomposed process advantage $A_{\text{decomposed},k}^{(i)}$, which is combined with the sparse outcome advantage $A_O^{(i)}$ to produce the final training signal.}
   \label{fig:pdcr_method}
\end{figure*}
Figure~\ref{fig:observation}-(d) (\textit{left}) illustrates this effect: the naive global normalization compresses and misaligns the \textcolor{VisualPerception}{\textbf{perception}} advantage distribution, which could lead to ill-scaled credit for these critical actions. In contrast, Figure~\ref{fig:observation}-(d) (\textit{right}) shows the effect of decomposed normalization. By normalizing each skill cluster independently, this mixture-induced signal degradation is mitigated. The resulting advantage distributions for both skills appear better scaled and aligned, aiming to provide a more stable and representative training signal. 

%% file: sec/5_method.tex
\section{Method: Perception-Decomposed Confidence Reward (PDCR)}
\label{sec:method}

Our method, Perception-Decomposed Confidence Reward (PDCR), directly addresses the core problem identified in Section~\ref{sec:challenge}: that a single, homogeneous reward is suboptimal for heterogeneous V-L skills. As illustrated in Figure~\ref{fig:pdcr_method}, PDCR operates in two stages. Specifically, it (i) performs \textbf{unsupervised skill decomposition} to automatically cluster steps into \textcolor{VisualPerception}{\textbf{visual perception}} and \textcolor{TextualReasoning}{\textbf{textual reasoning}} groups, and (ii) computes a \textbf{decomposed advantage} by normalizing confidence gains \textit{within} each respective cluster, ensuring a more stable, intra-cluster comparison.

\subsection{Unsupervised Skill Decomposition via Visual Dependence}
\label{sec:method_1}
To decompose the reward signal during training, we must categorize each reasoning step $h_k^{(i)}$ as \textcolor{VisualPerception}{\textbf{visual perception}} or \textcolor{TextualReasoning}{\textbf{textual reasoning}}. While our analysis in Section~\ref{sec:challenge} used externally-annotated labels (via GPT) to \textit{validate} the problem, such labels are not available in a practical training setup. Therefore, to achieve a scalable and self-contained decomposition, we introduce the \textbf{Visual Dependence Score}, $V_k^{(i)}$: a model-internal signal that quantifies a step's reliance on the visual input.

\begin{figure}[t!]
  \centering
   \includegraphics[width=1.0\linewidth]{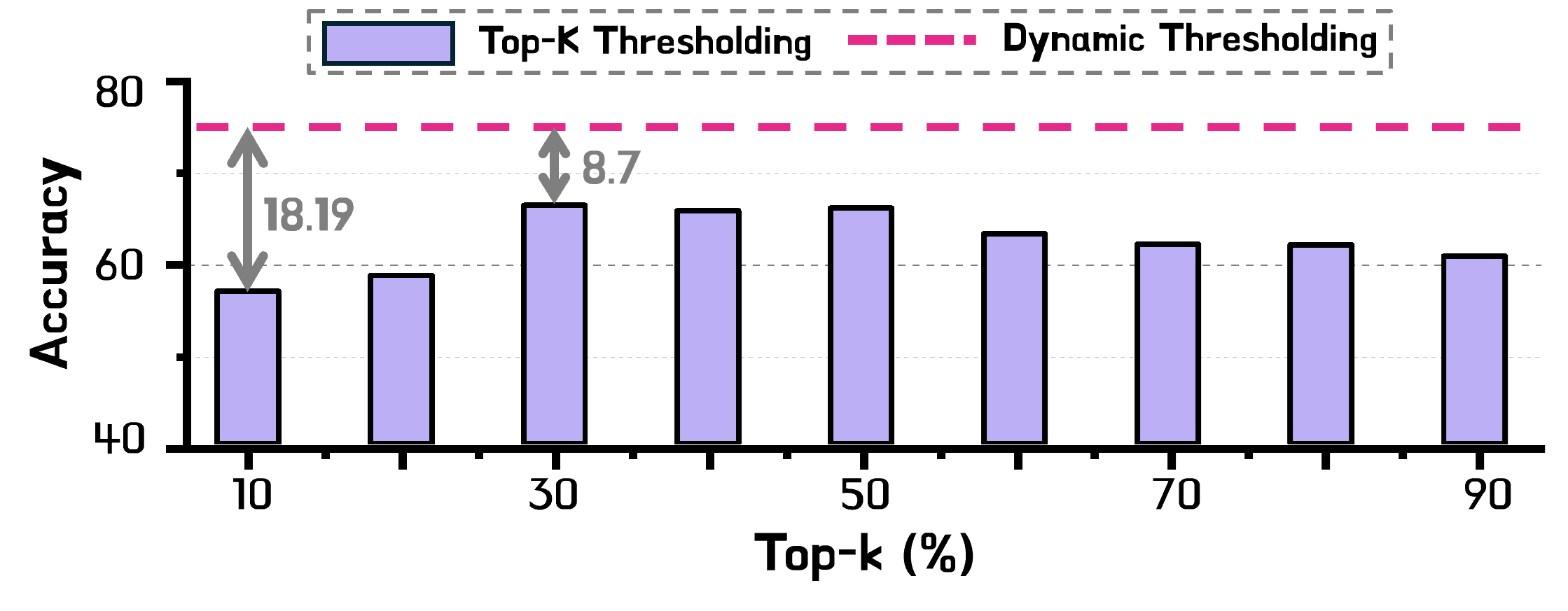}
\caption{Our dynamic thresholding (Otsu's method) is more accurate and robust at decomposing reasoning skills than a naive Top-K baseline. The Top-K method is highly sensitive to the $k$ hyperparameter, peaking at 30\%. Our parameter-free dynamic method significantly outperforms even the best Top-K, confirming it's a superior approach for skill decomposition.}
   \label{fig:thresholding}
\end{figure}

This score quantifies how much the generation of a specific step $h_k^{(i)}$ depended on the actual visual input $\mathbf{I}$, as opposed to a non-informative, blank (i.e., white) image $\mathbf{I}_{\text{white}}$. We define the standard log-probability of the step given the real image as $p_{k}^{(i)} = \log \pi_{\theta}(h_k^{(i)} | \mathbf{I}, \mathbf{q}, H_{<k}^{(i)})$ and the white-image log-probability as $p_{w,k}^{(i)} = \log \pi_{\theta}(h_k^{(i)} | \mathbf{I}_{\text{white}}, \mathbf{q}, H_{<k}^{(i)})$. The Visual Dependence Score $V_k^{(i)}$ is then the log-likelihood ratio of these two:
\begin{equation}
    \label{eq:vis_dependence}
    V_k^{(i)} = p_{k}^{(i)} - p_{w,k}^{(i)}.
\end{equation}

A high $V_k^{(i)}$ score indicates that the model relied heavily on the visual input $\mathbf{I}$ to generate the step $h_k^{(i)}$, marking it as a \textcolor{VisualPerception}{\textbf{visual perception}} step. A score near zero indicates the step was primarily driven by the text prefix $H_{<k}^{(i)}$, marking it as a \textcolor{TextualReasoning}{\textbf{textual reasoning}} step.

To separate steps, we dynamically find an optimal threshold $c^*$ that partitions the set of all visual dependence scores $\mathcal{V} = \{V_k^{(i)}\}_{(i,k) \in I}$. We determine $c^*$ using \textit{Otsu's method} \cite{otsu}, a classic and provably optimal 1D clustering algorithm. This method is widely used in image segmentation to find a threshold that best separates a pixel intensity into two classes (e.g., foreground and background). We apply it here to our 1D distribution of visual dependence scores to find the optimal split that minimizes the total intra-cluster variance.

First, let $\mathcal{V}_{\text{sorted}} = \{v_1, v_2, \dots, v_M\}$ be the set of all $M = |I|$ visual dependence scores, sorted in ascending order. We then iterate through all possible split points $k \in [1, M-1]$. Each $k$ partitions the sorted data into two clusters: $C_1(k) = \{v_1, \dots, v_k\}$ and $C_2(k) = \{v_{k+1}, \dots, v_M\}$.

For each potential split $k$, we compute the cluster means $\mu_1(k) = \frac{1}{k}\sum_{i=1}^{k} v_i$ and $\mu_2(k) = \frac{1}{M-k}\sum_{i=k+1}^{M} v_i$. We then calculate the total within-cluster sum of squares (SSE) for that split:
\begin{equation}
    \label{eq:sse}
    SSE(k) = \sum_{i=1}^{k} (v_i - \mu_1(k))^2 + \sum_{i=k+1}^{M} (v_i - \mu_2(k))^2.
\end{equation}
The optimal split point $k^*$ is the one that minimizes this SSE:
\begin{equation}
    \label{eq:argmin_sse}
    k^* = \underset{k}{\arg \min} \, SSE(k).
\end{equation}
The final threshold $c^*$ is set as the value of the score at this optimal split, $c^* = v_{k^*}$. 
This threshold $c^*$ partitions the set of all step indices $I$ into two disjoint subsets:
\begin{align}
    \mathcal{I}_{\text{visual}} &= \{ (i,k) \in I \mid V_k^{(i)} \ge c^* \}, \\
    \mathcal{I}_{\text{textual}} &= \{ (i,k) \in I \mid V_k^{(i)} < c^* \}.
\end{align}

\paragraph{Validating Dynamic Thresholding.} We validate this dynamic, data-driven threshold $c^*$ against a simpler \textit{top-K thresholding} baseline. This naive approach simply classifies a fixed percentage $k$ of steps with the highest $V_k^{(i)}$ scores as \textcolor{VisualPerception}{\textbf{visual perception}}. As shown in Figure~\ref{fig:thresholding}, we measure the classification accuracy of both methods against the ground-truth annotated labels from Section~\ref{sec:challenge}. The Top-K method (purple bars) is highly sensitive to the choice of $k$ and achieves a peak accuracy of 67.5\% at $k=30$. Our dynamic Otsu's method (pink dashed line) is parameter-free and significantly outperforms even the best Top-K baseline, achieving 76.2\% accuracy. This confirms that our dynamic approach is a more robust and accurate method for separating the two skill distributions.

\subsection{Perception-Decomposed Advantage Calculation}

With the step indices now decomposed, we compute a separate, normalized advantage for each skill cluster. We take the discounted returns $G_k^{(i)}$ (as defined in Section~\ref{sec:pacr}) and normalize them using min-max scaling, but only against other returns \textit{from the same cluster}.

As shown in Figure~\ref{fig:pdcr_method}, this yields two distinct process-level advantages. For visual steps, the advantage is:
\begin{equation}
    \label{eq:vis_advantage}
    A_{V,k}^{(i)} = \frac{G_{k}^{(i)} - \min\limits_{(j,k') \in \mathcal{I}_{\text{visual}}} G_{k'}^{(j)}}{\max\limits_{(j,k') \in \mathcal{I}_{\text{visual}}} G_{k'}^{(j)} - \min\limits_{(j,k') \in \mathcal{I}_{\text{visual}}} G_{k'}^{(j)}}.
\end{equation}
For textual steps, the advantage is:
\begin{equation}
    \label{eq:tex_advantage}
    A_{T,k}^{(i)} = \frac{G_{k}^{(i)} - \min\limits_{(j,k') \in \mathcal{I}_{\text{textual}}} G_{k'}^{(j)}}{\max\limits_{(j,k') \in \mathcal{I}_{\text{textual}}} G_{k'}^{(j)} - \min\limits_{(j,k') \in \mathcal{I}_{\text{textual}}} G_{k'}^{(j)}}.
\end{equation}
This decomposed normalization solves the ``mixture-induced signal degradation" problem. The gain of a perception step is now evaluated only against its peers, resulting in a stable, well-scaled reward signal for both skills.

\subsection{Final PDCR Advantage}

Finally, the total advantage $A_{total,k}^{(i)}$ for any given step is the weighted sum of the sparse outcome advantage $A_{O}^{(i)}$ (from Eq.~\ref{eq:sparse_advantage}) and its corresponding \textit{decomposed} process advantage $A_{\text{decomposed},k}^{(i)}$:
\begin{equation}
    \label{eq:pdcr_combined}
    A_{total,k}^{(i)} = \lambda_{O} A_{O}^{(i)} + \lambda_{P} A_{\text{decomposed},k}^{(i)},
\end{equation}
where $A_{\text{decomposed},k}^{(i)}$ is selected based on the step's cluster:
\begin{equation}
    \label{eq:pdcr_cases}
    A_{\text{decomposed},k}^{(i)} = \begin{cases} A_{V,k}^{(i)}, & \text{if } (i,k) \in \mathcal{I}_{\text{visual}} \\ A_{T,k}^{(i)}, & \text{if } (i,k) \in \mathcal{I}_{\text{textual}}. \end{cases}
\end{equation}
This final advantage $A_{total,k}^{(i)}$ is then applied to all tokens comprising the step $h_k^{(i)}$ during the policy update.

\begin{table*}[h!]
    \centering
\caption{\textbf{Main results on V-L reasoning benchmarks.} We report the accuracy across seven evaluation benchmarks. We compare our method, \textbf{PDCR (ours)}, against strong baselines, including sparse-reward (GRPO), stabilized (DAPO), and naive dense-reward (PACR) methods. The \textbf{best} score in each column is in bold, and the \underline{second-best} is underlined.}
    \vskip -0.1in
    \resizebox{1.0\textwidth}{!}{
    \begin{tabular}{l|cccc|ccc|c}
    \Xhline{4\arrayrulewidth}
    \textbf{Base model + Method} &
    \multicolumn{4}{c|}{\textbf{General Visual Understanding}} &
    \multicolumn{3}{c|}{\textbf{Visual Math \& Hallucination}} &
     \\
    \cline{2-9}
    & MMMU-Pro & MMMU & RealWorldQA & VisNumBench
    & MathVerse & MathVision & HallusionBench
    & \textbf{Average} \\
    \Xhline{2\arrayrulewidth}
        Visionary-R1 (3B) by \cite{xia2025visionary} & 27.4 & 30.6 & 56.9 & 10.0 & 45.0 & 36.5 & 30.0 & 33.8 \\
        Perception-R1 (7B) by \cite{xiao2025advancing} & 36.8 & 40.9 & 69.4 & 15.9 & 52.1 & 35.7 & 65.4 & 45.2 \\
        Vision-R1 (7B) by \cite{huang2025vision} & 34.9 & 42.8 & 60.1 & 33.0 & 57.3 & 51.2 & 32.2 & 44.5 \\
        \Xhline{4\arrayrulewidth}
       \rowcolor{table-yellow!60}\textit{Backbone model: Qwen2.5-VL-3B} \\
        \hspace{0.2em} Zero-shot Inference & 30.5 & 24.0 & 56.7 & 15.4 & 43.7 & 38.6 & 45.5 & 36.3 \\
        \hspace{0.2em} GRPO \cite{grpo} & 32.8 & 40.4 & 56.6 & 29.9 & 43.5 & \underline{39.4} & \underline{62.6} & 43.6\\
        \hspace{0.2em} DAPO \cite{dapo} & 32.6 & 41.0 & 57.5 & 31.8 & 44.0 & 39.1 & \textbf{62.7} & 44.1 \\
        \hspace{0.2em} PACR \cite{yoon2025pacr} & \textbf{33.4} & \underline{41.2} & 57.6 & \underline{32.9} & \underline{45.2} & 39.0 & 61.5 & \underline{44.4} \\
        \rowcolor{table-green!40} \hspace{0.2em} \textbf{PDCR (ours)} & \underline{33.3} & \textbf{43.2} & \textbf{59.3} & \textbf{33.3} & \textbf{45.6} & \textbf{40.1} & 61.8 & \textbf{45.2}\\
        \Xhline{3\arrayrulewidth}
       \rowcolor{table-yellow!60}\textit{Backbone model: Qwen2.5-VL-7B} \\
        \hspace{0.2em} Zero-shot Inference & 33.2 & 33.3 & 65.5 & 12.6 & 52.9 & 40.3 & 52.2 & 41.4 \\
        \hspace{0.2em} GRPO \cite{grpo} & \underline{41.9} & 50.7 & 65.6 & \textbf{41.5} & 53.2 & 41.3 & 66.6 & 51.5\\
        \hspace{0.2em} DAPO \cite{dapo} & 41.8 & \underline{51.0} & 66.4 & 40.0 & 53.9 & 44.2 & 67.0 & 52.0  \\
        \hspace{0.2em} PACR \cite{yoon2025pacr} & 41.5 & 50.5 & \underline{66.9} & 40.1 & \underline{54.3} & \underline{44.7} & \underline{67.6} & \underline{52.2}\\
        \rowcolor{table-green!40} \hspace{0.2em} \textbf{PDCR (ours)} & \textbf{42.5} & \textbf{51.5} & \textbf{67.3} & \underline{41.1} & \textbf{55.0} & \textbf{44.8} & \textbf{68.4} & \textbf{52.9} \\ 
        \Xhline{4\arrayrulewidth}
    \end{tabular}
    }
    \label{tab:main_result}
\end{table*}

\begin{table*}[h!]
    \centering
\caption{\textbf{Ablation study on the efficacy of skill decomposition.} We compare our full PDCR method against a \textit{Random Decomposition} baseline. This baseline randomly assigns steps to the visual or textual cluster before applying decomposed normalization. The clear performance gap validates that our data-driven \textit{Visual Dependence Score} provides a meaningful decomposition, and that simply decomposing the reward is not sufficient.}
    \vskip -0.1in
    \resizebox{1.0\textwidth}{!}{
    \begin{tabular}{l|cccc|ccc|c}
    \Xhline{4\arrayrulewidth}
    \textbf{Base model + Method} &
    \multicolumn{4}{c|}{\textbf{General Visual Understanding}} &
    \multicolumn{3}{c|}{\textbf{Visual Math \& Hallucination}} &
     \\
    \cline{2-9}
    & MMMU-Pro & MMMU & RealWorldQA & VisNumBench
    & MathVerse & MathVision & HallusionBench
    & \textbf{Average} \\
    \Xhline{2\arrayrulewidth}
       \rowcolor{table-yellow!60}\textit{Backbone model: Qwen2.5-VL-3B} \\
        \rowcolor{table-green!40} \hspace{0.2em} \textbf{PDCR (ours)} & \textbf{33.3} & \textbf{43.2} & \textbf{59.3} & \textbf{33.3} & \textbf{45.6} & \textbf{40.1} & 61.8 & \textbf{45.2} \\
        \hspace{0.6em}\hspace{0.2em} $\rightarrow$ Random Decomposition & 33.1 & 41.0 & 57.9 & 31.0 & 44.5 & 39.3 & \textbf{61.9} & 44.1 \\
        \Xhline{3\arrayrulewidth}
       \rowcolor{table-yellow!60}\textit{Backbone model: Qwen2.5-VL-7B} \\
        \rowcolor{table-green!40} \hspace{0.2em} \textbf{PDCR (ours)} & \textbf{42.5} & \textbf{51.5} & \textbf{67.3} & \textbf{41.1} & \textbf{55.0} & \textbf{44.8} & \textbf{68.4} & \textbf{52.9} \\ 
        \hspace{0.6em}\hspace{0.2em} $\rightarrow$ Random Decomposition & 41.6 & 51.2 & 67.1 & 40.6 & 54.6 & 43.7 & 67.2 & 52.3 \\
        \Xhline{4\arrayrulewidth}
    \end{tabular}
    }
    \label{tab:random}
\end{table*}

%% file: sec/6_experiments.tex
\section{Experimental Setup}
\subsection{Models and Datasets}
Our experiments utilize two open-source VLMs: \textbf{Qwen2.5-VL-3B-Instruct} and \textbf{Qwen2.5-VL-7B-Instruct}. We train all models on the \textbf{Vision-SR1} dataset \cite{sr1}. Vision-SR1 is a diverse training corpus of approximately 47K examples, all of which include verifiable answers suitable for GRPO training. The dataset is specifically balanced to cover three key reasoning domains: Mathematical Reasoning, Commonsense Knowledge, and General Visual Understanding.
\subsection{Evaluation Benchmarks}
\paragraph{General Visual Understanding.}
We evaluate on \textbf{MMMU} \cite{mmmu}, a 11.5K-question college-level benchmark for cross-disciplinary knowledge; its more rigorous successor, \textbf{MMMU-Pro} \cite{mmmu_pro}, which increases the difficulty; \textbf{RealWorldQA} \cite{RealWorldQA2024}, ~700 real-world images with spatially-grounded questions; and \textbf{VisNumBench} \cite{weng2025visnumbench}, a ~1.9K question benchmark targeting numerical estimation and counting.

\paragraph{Multimodal Mathematical Reasoning.}
We use two specialized benchmarks: \textbf{MathVerse} \cite{zhang2024mathverse}, which evaluates Chain-of-Thought (CoT) correctness on 2.6K diagram-centric problems; and \textbf{MATH-Vision} \cite{mathvision}, ~3K competition-grade problems across 16 disciplines.

\paragraph{Hallucination Diagnosis.}
We use \textbf{HallusionBench} \cite{guan2024hallusionbench}, an image-context reasoning benchmark specifically designed to test a model's hallucination and visual illusion errors.
\subsection{Baselines}
\label{sec:baselines}
We compare our proposed PDCR against several key baselines on both model backbones. For a direct and fair comparison, we train using identical system prompts and hyperparameters (details in Appendix~\ref{app:implementation_detail}): \textbf{GRPO} \cite{grpo}; \textbf{DAPO} \cite{dapo}, which solves the vanishing advantage problem of GRPO (for this baseline, we add only dynamic sampling to standard GRPO, without incorporating other improvements from DAPO such as a higher clip rate.); and \textbf{PACR} \cite{yoon2025pacr}, which applies a naive, globally-normalized ground-truth confidence growth process reward. 
\section{Experimental Results and Ablations}
\label{sec:exp}

\begin{figure*}[t!]
  \centering
   \includegraphics[width=1.0\linewidth]{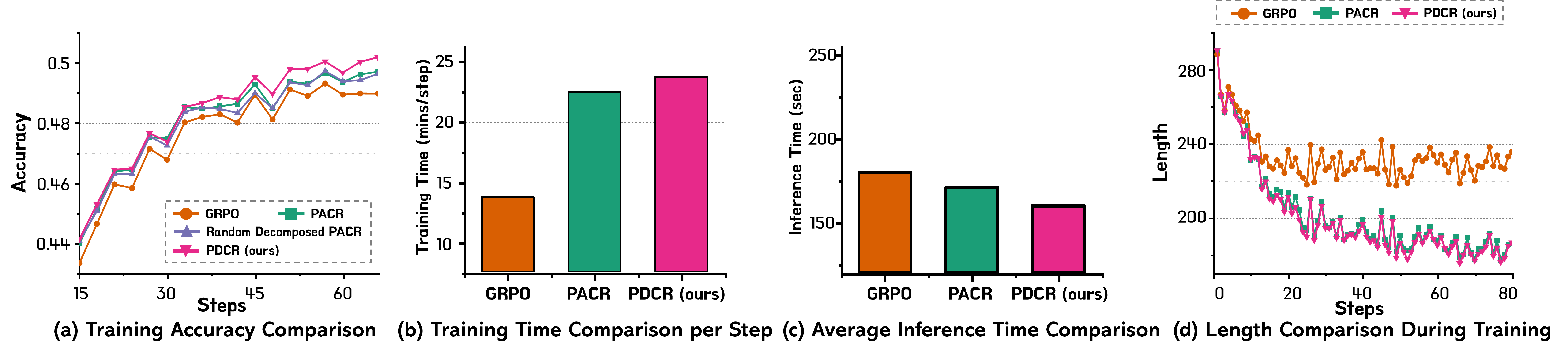}
   \vspace{-0.7cm}
\caption{Training dynamics, cost, and efficiency comparison.
\textbf{(a)} Our PDCR method (pink) converges faster and to a higher final accuracy than all baselines.
\textbf{(b)} This improved performance comes at an expected, higher computational cost per training step.
\textbf{(c) \& (d)} This training cost is offset by a significant gain in inference-time efficiency (c). As shown in (d), this efficiency gain is driven by the fact that both PDCR and PACR learn to produce substantially shorter and more concise reasoning traces over time compared to GRPO.}
   \label{fig:train_curve}
\end{figure*}

\paragraph{Main Results.} Table~\ref{tab:main_result} presents the main experimental results on both the Qwen2.5-VL-3B and Qwen2.5-VL-7B backbones. Across both model scales, our PDCR framework outperforms all baselines.

On the 7B model, PDCR achieves a final average score of 52.9, surpassing the next-best method, PACR (52.2), as well as the GRPO baseline (51.5). This demonstrates the clear benefit of our decomposed reward. Notably, PDCR shows consistent improvements over the naive, globally-normalized PACR, particularly on complex understanding tasks like MMMU-Pro (42.5 vs. 41.5) and MMMU (51.5 vs. 50.5), as well as on MathVerse (55.0 vs. 54.3).
This performance gain is consistent on the 3B model, where PDCR (45.2) achieves the highest average score, outperforming the naive PACR (44.4) and the sparse GRPO (43.6).

We also note a key distinction from GRPO. A well-known issue of GRPO is the \textbf{vanishing advantage problem}: if all $N$ trajectories in a group are either correct or incorrect, the outcome advantage $A_{O}^{(i)}$ becomes zero for that sample, providing no learning signal. The DAPO baseline is designed to mitigate this by re-sampling until a mixed-outcome group is obtained. Our PDCR, by definition, solves this as well, since its process-level reward $A_{P,k}^{(i)}$ is non-zero. A key question, therefore, is whether PDCR's gains are merely from stabilization. However, our results show that \textit{PDCR also consistently outperforms DAPO} (e.g., 52.9 vs. 52.0 and 45.2 vs. 44.1). This comparison validates that our perception-decomposed reward provides a meaningful, high-quality training signal, and its contribution goes beyond just mitigating the vanishing advantage problem of GRPO. 

\paragraph{Efficacy of Skill Decomposition.}

To isolate the impact of our \textit{unsupervised skill decomposition} (Section~\ref{sec:method_1}), we compare with a \textbf{Random Decomposition} baseline.

This baseline follows the same PDCR framework, including the decomposed advantage calculation (Eq.~\ref{eq:vis_advantage} and \ref{eq:tex_advantage}). However, instead of using our Visual Dependence Score ($V_k^{(i)}$) and dynamic thresholding to cluster the steps, it simply assigns each step to the $\mathcal{I}_{\text{visual}}$ or $\mathcal{I}_{\text{textual}}$ cluster at random. 

The results are presented in Table \ref{tab:random}. On the 7B backbone, the Random Decomposition baseline (52.3) performs significantly worse than our full PDCR method (52.9) and only slightly better than the naive PACR (52.2). This demonstrates that simply decomposing the reward is not enough. The performance gain of our method comes from \textit{correctly} identifying and separating the heterogeneous skills. By using a data-driven, unsupervised clustering method, our PDCR framework provides a meaningful decomposition that effectively trains both perception and reasoning.

\paragraph{Training Dynamics.}
We analyze the training dynamics in Figure \ref{fig:train_curve}-(a). The plot, which tracks average accuracy over training steps, shows that our PDCR method consistently outperforms all baselines throughout the entire training process. While the sparse GRPO baseline shows the slowest improvement, the naive PACR and our Random Decomposition baseline perform very similarly, reinforcing our earlier finding that a random decomposition offers minimal benefit. 

\paragraph{Training Cost and Efficiency.}
We analyze computational overhead in Figure~\ref{fig:train_curve}-(b). PACR incurs a $\sim$1.5x cost increase over GRPO, as it requires an extra forward pass to compute $c_k$ (Eq.~\ref{eq:step_confidence}). Our PDCR adds a minimal further overhead, requiring one additional forward pass per \textit{trajectory} to compute the $V_k$ (Eq.~\ref{eq:vis_dependence}), compared to PACR. Crucially, this training cost is offset by a significant gain in generation efficiency as shown in Figure \ref{fig:train_curve}-(c). This gain is driven by the fact that both PACR and our PDCR learn to produce \textit{shorter, more concise reasoning traces} over time, as shown in Figure~\ref{fig:train_curve}-(d). 

%% file: sec/7_conclusion.tex
\section{Conclusion}
\label{sec:conclusion}
We introduced PDCR to address the challenge of heterogeneous skills in Vision-Language reasoning. By clustering steps via a model-internal Visual Dependence Score, PDCR decomposes the confidence growth reward in an unsupervised manner. This intra-cluster normalization resolves mixture-induced signal degradation, providing stable advantages for both perception and reasoning. Ultimately, PDCR improves strong RLVR baselines, such as GRPO and PACR, using a self-contained signal.

%% file: sec/X_suppl.tex
\clearpage
\setcounter{page}{1}
\maketitlesupplementary

\addcontentsline{toc}{section}{Appendix} 

\startcontents[appendix]
\printcontents[appendix]{}{1}{\subsection*{Appendix Contents}}

\vspace{2em} 
\noindent\rule{\linewidth}{0.8pt} 
\vspace{2em}
\section{Broader Impact}
\label{sec:broader_impact}
This work introduces a framework for improving the reasoning capabilities of multimodal Large Language Models. By leveraging the model's intrinsic confidence dynamics, our method provides fine-grained, step-level supervision, and decomposes this signal to align with the heterogeneous skills of perception and reasoning. This is achieved without the overhead of training separate reward models or requiring manual data annotation. 

\section{Ethics Statement}
\label{sec:ethics}
This research strictly adheres to academic integrity standards, ensuring all prior work is properly cited and acknowledged. Furthermore, our experiments utilize only publicly available datasets and do not involve the collection of sensitive or personally identifiable information.
\clearpage

\newcommand{\pdcr}[1]{\textcolor{PDCRColor}{#1}}

\section{Training Procedure Pseudocode}
We outline our Perception-Decomposed Confidence Reward (PDCR) training procedure in Algorithm~\ref{alg:pdcr}. This pseudocode provides a step-by-step specification of the method summarized in Section~\ref{sec:method}. The \pdcr{highlighted lines} indicate the additional processing steps introduced in our proposed PDCR compared to PACR \cite{yoon2025pacr}. 

\input{pseudocode/pdcr}
\clearpage
\section{Experimental Results on Additional Model Backbone}
\label{app:add_exp}

We further evaluate PDCR on the recently released \href{https://huggingface.co/Qwen/Qwen3-VL-8B-Instruct}{Qwen3-VL-8B-Instruct} (implementation details are outlined in Appendix \ref{app:implementation_detail}).
As shown in Table \ref{tab:qwen3}, PDCR demonstrates generalization to this stronger backbone, achieving a final average score of \textbf{59.1}. This performance outperforms the sparse GRPO baseline (58.3, \textbf{+0.8\%}) and also improves upon the naive dense-reward baseline, PACR (58.5, \textbf{+0.6\%}).

\begin{table*}[t!]
    \centering
\caption{\textbf{Experimental results on the Qwen3-VL-8B-Instruct backbone.} We report the accuracy across seven evaluation benchmarks. We compare our method, \textbf{PDCR (ours)}, against strong baselines, including sparse-reward (GRPO), stabilized (DAPO), and naive dense-reward (PACR) methods. The \textbf{best} score in each column is in bold, and the \underline{second-best} is underlined.}
    \vskip -0.1in
    \resizebox{1.0\textwidth}{!}{
    \begin{tabular}{l|cccc|ccc|c}
    \Xhline{4\arrayrulewidth}
    \textbf{Base model + Method} &
    \multicolumn{4}{c|}{\textbf{General Visual Understanding}} &
    \multicolumn{3}{c|}{\textbf{Visual Math \& Hallucination}} &
     \\
    \cline{2-9}
    & MMMU-Pro & MMMU & RealWorldQA & VisNumBench
    & MathVerse & MathVision & HallusionBench
    & \textbf{Average} \\
    \Xhline{2\arrayrulewidth}
       \rowcolor{table-yellow!60}\textit{Backbone model: Qwen3-VL-8B} \\
        \hspace{0.2em} Zero-shot Inference & 49.1 & 55.2 & 70.1 & 29.1 & 68.1 & 48.3 & 74.9 & 56.4\\
        \hspace{0.2em} GRPO \cite{grpo} & 
        49.6 & 55.0 & \underline{71.2} & \underline{37.2} & 69.3 & 49.9 & 75.6 & 58.3 \\
        \hspace{0.2em} DAPO \cite{dapo} & 50.0 & 55.5 & \textbf{71.3} & 37.0 & 69.2 & 50.3 & \underline{75.8} & 58.4 \\
        \hspace{0.2em} PACR \cite{yoon2025pacr} & \underline{49.9} & \underline{56.1} & 70.6 & 37.1 & \underline{69.9} & \underline{50.4} & 75.5 & \underline{58.5} \\
        \rowcolor{table-green!40} \hspace{0.2em} \textbf{PDCR (ours)} & \textbf{50.7} & \textbf{57.1} & 70.7 & \textbf{37.6} & \textbf{70.6} & \textbf{51.0} & \textbf{76.0} & \textbf{59.1}\\
        \Xhline{4\arrayrulewidth}
    \end{tabular}
    }
    \label{tab:qwen3}
\end{table*}

\section{Segmentation Detail}
\label{app:segmentation}
A prerequisite for a process-based reward framework is the segmentation of the reasoning trajectory $\tau^{(i)}$ into a discrete sequence of steps $\{h^{(i)}_k\}_{k=1}^{K_i}$. The step is the fundamental unit to which a reward or advantage is assigned.

Previous work in process-reward modeling has adopted several strategies to define this unit:

\begin{itemize}
\item \textbf{Supervised Formatting via SFT:} Some approaches utilize Supervised Fine-Tuning (SFT) to enforce a highly structured output format, such as training the model to generate explicit step delimiters (e.g., ``Step 1:'', ``Step 2:''). While this simplifies segmentation, it necessitates an SFT stage to instill these formatting constraints. As detailed in our implementation (Appendix~\ref{app:implementation_detail}), our training adopts a R1-style pipeline \cite{deepseekr1} that bypasses SFT to initiate reasoning training directly via RLVR. Consequently, strategies relying on pre-learned formatting are not applicable to our setting.

    \item \textbf{Confidence-based Segmentation:} Some approaches leverage intrinsic model signals to dynamically segment reasoning trajectories. For instance, \citet{liu2025adaptivestep} proposes, a method that divides reasoning steps based on the model's confidence in predicting the next token. This approach aims to identify true decision points by observing how the model's uncertainty changes, avoiding arbitrary segmentation based on fixed lengths or rule-based heuristics that may not align with the model's internal reasoning flow.

    \item \textbf{Rule-Based Segmentation:} A more common, lightweight, and model-agnostic approach is to use simple heuristics to parse the generated text. This is the strategy adopted by the original PACR framework \cite{yoon2025pacr} and other process-reward based methods \cite{yang2025beyond,zeng2025versaprm,lee2025rethinking}. The most common heuristic is to define a new step at each newline character (\texttt{\textbackslash n}), or after sentence-terminating punctuation (e.g., a period followed by a space). \textbf{In this work, we adopt this strategy due to its simplicity and to ensure a fair comparison with PACR \cite{yoon2025pacr}.}
\end{itemize}

\clearpage

\section{Label Acquisition for Skill Analysis}
\label{app:gpt_label}

To empirically validate the heterogeneous nature of V-L reasoning (\texttt{\textbf{[Observation 1]}} in Section \ref{sec:challenge}) and the effectiveness of our unsupervised skill decomposition (Section \ref{sec:method_1}), we required a set of ground truth skill labels. Since no existing dataset provides step-level distinctions between perception and reasoning, we constructed a high-quality annotated set using a state-of-the-art Large Language Model as a programmatic annotator.

\subsection{Annotation Setup}
\label{app:annotation}
We sampled 10,000 reasoning trajectories generated by the \texttt{Qwen2.5-VL-7B-Instruct} model from our validation set. Each trajectory was first segmented into discrete steps following Appendix~\ref{app:segmentation}. Subsequently, we utilized \texttt{gpt-5-2025-08-07} to classify every individual step within these trajectories.
To ensure the labels captured true visual dependency rather than just semantic relevance, we designed a rigorous prompt with a specific exclusion criterion. The model was provided with the full context (Image, Question, Full Reasoning Trace) and the specific Candidate Step, and instructed to output a boolean classification in strict JSON format. The exact system instruction provided to the annotator is as follows:

\begin{tcolorbox}
[colback=gray!5!white,colframe=gray!75!black,title=System Instruction for Skill Annotation]
\label{box:prompt}
\footnotesize
\textbf{Role:} You are an expert annotator. Decide whether a SPECIFIC step in a chain-of-thought is a ``visual perception" step.

\textbf{Definition:} A visual perception step should primarily involve reading or describing visual content from the provided image (e.g., recognizing objects, numbers, colors, relative positions, text in the image, bars in a chart, etc.). Steps that perform logical deductions, arithmetic, or generic reasoning without referencing observed visual features are NOT visual-perception steps.

\textbf{Exclusion Criterion:} If a step refers to visual details that could already be inferred from earlier reasoning (i.e., it does not require directly inspecting the image again), classify that step as \textbf{NOT} visual perception.

\textbf{Output Format:} Return STRICT JSON with keys exactly: 
\begin{itemize}
    \item \texttt{is\_visual\_perception} (boolean)
    \item \texttt{reason} (string)
\end{itemize}
Do not include extra keys or text outside JSON.

\hrulefill

\textbf{Input Template:} \\
Question: \texttt{\{question\}} \\
Full thinking process (all steps concatenated): \texttt{\{full\_process\}} \\
Candidate step to classify: \texttt{\{step\_text\}}

Now output JSON only.
\end{tcolorbox}

\paragraph{Distinguishing Perception from Contextual Recall.}
A critical challenge in defining these labels is separating \textit{active perception} from \textit{visual memory}. As noted in the \textbf{Exclusion Criterion} above, a reasoning step might reference a visual object without actually looking at the image, relying instead on facts established in previous steps. We explicitly instructed the model to classify such instances as \textcolor{TextualReasoning}{\textbf{Textual Reasoning}}, ensuring that the \textcolor{VisualPerception}{\textbf{Visual Perception}} label is reserved strictly for steps that extract \textit{new} information from the raw visual signal.

\subsection{Validation of Label Quality}
To verify the reliability of this automated annotation, we performed a rigorous inter-annotator agreement study:

\begin{enumerate}
    \item \textbf{Human Inter-Annotator Agreement:} Two human experts independently annotated a random subset of 100 steps. They achieved a Cohen's Kappa of $\kappa=0.82$, indicating that the binary distinction between perception and reasoning is well-defined and unambiguous to humans.
    \item \textbf{Model-Human Alignment:} We compared the primary \texttt{gpt-5} annotations against the human consensus on the same subset. The model achieved a Kappa score of $\kappa=0.79$ (Table~\ref{tab:kappa_scores}). This high alignment confirms that the model effectively acts as a reliable proxy for human judgment, correctly adhering to the strict definitions provided in the prompt.
\end{enumerate}

\begin{table}[h]
    \centering
    \small
    \caption{\textbf{Inter-Annotator Agreement Scores.} The strong correlation ($\kappa > 0.75$) validates that the labels are reliable.}
    \label{tab:kappa_scores}
    \begin{tabular}{llc}
        \toprule
        \textbf{Comparison Pair} & \textbf{Metric Interpretation} & \textbf{Cohen's $\kappa$} \\
        \midrule
        Human Expert 1 vs. Expert 2 & Task Definition Quality & 0.82 \\
        GPT-5 vs. Human Consensus & Proxy Reliability & 0.79 \\
        \bottomrule
    \end{tabular}
\end{table}
\clearpage

\subsection{Qualitative Examples of Skill Decomposition}
\label{app:qualitative}
To further illustrate the functional distinction between visual perception and textual reasoning (discussed as Observation 1 in the main text), we provide a set of representative reasoning steps in Table~\ref{tab:qual_examples}. Specifically, we show examples of individual steps that we labeled as either \textcolor{VisualPerception}{\textbf{Visual Perception}} or \textcolor{TextualReasoning}{\textbf{Textual Reasoning}} based on the annotation process described in Appendix~\ref{app:annotation}. As shown below, perception steps are characterized by explicit grounding in the image, whereas reasoning steps rely on logical deduction or internal knowledge.

\begin{longtable}{@{}p{0.03\textwidth}p{0.92\textwidth}@{}}
\caption{\textbf{Qualitative Examples of Skill Decomposition.} We show examples of steps that we labeled as \textcolor{VisualPerception}{\textbf{Visual Perception}} or \textcolor{TextualReasoning}{\textbf{Textual Reasoning}}. This separation illustrates the heterogeneous nature of the V-L reasoning task.}
\label{tab:qual_examples} \\
\toprule
\multicolumn{2}{l}{\textbf{Sample Step Content}} \\
\midrule
\endfirsthead

\toprule
\multicolumn{2}{l}{\textbf{Sample Step Content}} \\
\midrule
\endhead

\bottomrule
\endlastfoot

\multicolumn{2}{c}{\textbf{\textsc{Examples of Visual Perception Steps}}} \\
\multicolumn{2}{c}{\textit{(Extracting attributes, counting, reading text, identifying coordinates)}} \\
\midrule

$\bullet$ & The image shows a group of people in red and white attire that resembles soccer jerseys, who appear to be in distress or disappointment. \\
$\bullet$ & The use of impasto technique, where thickly applied paint creates texture, is also evident. \\
$\bullet$ & The label `G' is positioned in the middle of the solar system, closer to the sun than Venus and further from the sun than Earth. \\
$\bullet$ & The individual appears to be actively paddling and navigating the river, which suggests a sense of engagement and energy. \\
$\bullet$ & Looking at the diagram, the area that is light gray and surrounds the darker gray area (umbra) is the penumbra. \\
$\bullet$ & The diagram shows a plant cell with various labeled parts. \\
$\bullet$ & This person seems to be looking at a smaller, darker figure that is separate from the main silhouette, suggesting a sense of isolation or separation. \\
$\bullet$ & The image shows a group of players standing in front of a goalpost, and there is a ball present on the field. \\
$\bullet$ & The Moon is at stage 1, and the diagram indicates that it is at the position labeled as ``Midnight." Therefore, the time of day at stage 1 is midnight. \\

\midrule
\addlinespace[0.5em]
\multicolumn{2}{c}{\textbf{\textsc{Examples of Textual Reasoning Steps}}} \\
\multicolumn{2}{c}{\textit{(Deduction, arithmetic, applying external knowledge, conclusion drawing)}} \\
\midrule

$\bullet$ & Since the question asks for the total sum of the values, we need to add the value of the blue bar (45) to the value of the red bar (30), resulting in $45 + 30 = 75$. \\
$\bullet$ & Antarctica is the only continent in the world that is entirely below the Antarctic Circle, making it stand out as the only one often highlighted in this way on world maps. \\
$\bullet$ & Substituting these into the formula for the average value, we get: \\
$\bullet$ & Combining these observations, the time shown on the clock is 5:30. \\
$\bullet$ & To find the total area of the composite shape, we must calculate the area of the square and add it to the area of the triangle. \\
$\bullet$ & Therefore, based on the architectural style and location, the most likely location of the image is Xi'an. \\
$\bullet$ & In nature, this type of relationship is known as mutualism, which is a form of symbiotic relationship. \\
$\bullet$ & Applying the Pythagorean theorem to the right-angled triangle formed by the ladder and the wall, we can solve for the unknown length of the ladder. \\
$\bullet$ & Thus, we can infer that the vehicle entering the intersection has committed a traffic violation. \\
\end{longtable}

\clearpage
\section{Implementation Details}
\label{app:implementation_detail}

\subsection{Training Framework and Hyperparameters}
We perform all experiments using the \href{https://github.com/hiyouga/easyr1}{\textbf{EasyR1}} framework. Consistent with the R1-Zero style training \cite{deepseekr1}, we apply Reinforcement Learning with Verifiable Rewards (RLVR) directly on the base model, bypassing any Supervised Fine-Tuning (SFT) stage. This ensures that the reasoning behaviors we observe are genuine results of the reinforcement learning process, rather than artifacts inherited from supervised fine-tuning.

All models (including baselines) are trained on the \texttt{Vision-SR1-47K} dataset for 1 epoch. To ensure a strictly fair comparison, we maintain a consistent configuration across all methods (GRPO \cite{grpo}, DAPO \cite{dapo}, PACR \cite{yoon2025pacr}, and our PDCR). Common hyperparameters for the optimizer, RL process, and evaluation are detailed in Table~\ref{tab:hype}. All models were trained on a single node with 8 $\times$ NVIDIA A100 80GB GPUs.

\paragraph{Format reward.}
Given that our training pipeline bypasses SFT, the model does not inherently possess the structural priors to format its reasoning steps. To address this, we incorporate a format reward alongside the outcome accuracy reward. A trajectory is assigned a format score $R_{\text{fmt}}$ based on the valid enclosure of the reasoning process within \texttt{<think>} and \texttt{</think>} tags, and the final answer within \texttt{\textbackslash boxed\{\}}. Specifically, we augment the outcome reward $R^{(i)}$ (Eq.~\ref{eq:sparse_reward}) with a fixed bonus of $R_{\text{fmt}}=0.1$ for compliant trajectories, while assigning $R_{\text{fmt}}=0$ to invalid ones.

\paragraph{Details on DAPO baseline.}
For the DAPO baseline \cite{dapo}, we adopted a specific configuration to isolate the impact of reward stability. The full DAPO framework originally introduces three modifications to GRPO: (1) \textit{Clipping} to promote diversity and prevent entropy collapse, (2) \textit{Dynamic Sampling} to solve the vanishing advantage problem (where a batch has 0 variance in rewards), and (3) a \textit{Token-Level Policy Gradient} loss to mitigate length bias.

As discussed in Section~\ref{sec:exp}, our primary goal in comparing with DAPO is to determine if PDCR's performance gains are merely a result of stabilizing the training signal against the vanishing advantage problem. Therefore, our implementation of the DAPO baseline specifically applies \textit{only} Dynamic Sampling on top of the standard GRPO framework. This controlled configuration allows for a direct, isolated comparison regarding reward signal stability.

\begin{table}[h!]
\small
\centering
\caption{\textbf{Key hyperparameters} standard to the RLVR framework in EasyR1 library \cite{easyr1} used for training and evaluation.}
\begin{tabular}{ll}
\hline
\textbf{Hyperparameter} & \textbf{Value} \\
\hline
\rowcolor{gray!20} \multicolumn{2}{l}{\textit{Common Hyperparameters}} \\
Optimizer           & AdamW \cite{adamw} \\
Learning Rate       & 1e-6 \\
Weight Decay & 1e-2 \\
LR Schedule         & Constant \\
Epochs              & 1 \\
Freeze Vision Encoder & False \\
Global Batch Size    & 128 \\
Rollout Batch Size   & 512 \\
Rollout Size  & 8 \\
Rollout Temperature & 1.0 \\
Rollout Top-p        & 0.99 \\
Max Input Prompt Length & 12800 \\
Max Response Length  & 2048 \\
Max Pixels & 4194304 \\
Min Pixels & 262144 \\
Use Kl Loss & True \\
KL Coef & 1e-2 \\

\hline
\rowcolor{gray!20} \multicolumn{2}{l}{\textit{PACR \& PDCR Specific}} \\
$\lambda_O$ & 0.7 \\
$\lambda_P$ & 0.3 \\
\hline
\rowcolor{gray!20} \multicolumn{2}{l}{\textit{Evaluation Generation}} \\
Temperature      & 0.0 \\
Max New Tokens   & 2048 \\
\hline
\end{tabular}
\label{tab:hype}
\end{table}

\subsection{Prompt Template for Training and Inference}
\label{app:templates}

To ensure a fair comparison, we use the same system prompt template across GRPO, DAPO, PACR, and our PDCR method as follows:

\definecolor{rliableblue}{HTML}{77AADD}
\newcommand{\placeholder}[1]{\textcolor{red}{\{#1\}}}

\tcbset{
  reasoningbox/.style={
    enhanced,
    colback=white,
    colframe=rliableblue!80!black,
    colbacktitle=rliableblue!80!black,
    coltitle=white,
    fonttitle=\bfseries,
    boxrule=0.7pt,
    arc=2mm,
    left=3mm,
    right=3mm,
    top=2mm,
    bottom=2mm,
    boxsep=1.5pt
  }
}
\begin{tcolorbox}[reasoningbox,title=Reasoning Template for GRPO\text{,} DAPO\text{,} PACR\text{,} and PDCR]
\textbf{SYSTEM:} \\
You are a helpful assistant. \\[0.5em]

\textbf{USER:} \\
\placeholder{image} \\
\placeholder{question} \\[0.8em]

You FIRST analyze the image/video and think about the detailed, step-by-step reasoning process as an internal monologue and then provide the final answer. The reasoning process MUST BE enclosed within \textless think\textgreater\ \textless/think\textgreater\ tags. The final answer MUST BE put in \textbackslash boxed\{\}.
\end{tcolorbox}

\vspace{4em}

\clearpage
\section{Ablation Study on Visual Dependence Calculation for Skill Decomposition}
\label{app:ablation_masking}

\begin{figure}[h]
    \centering
    \includegraphics[width=1.0\linewidth]{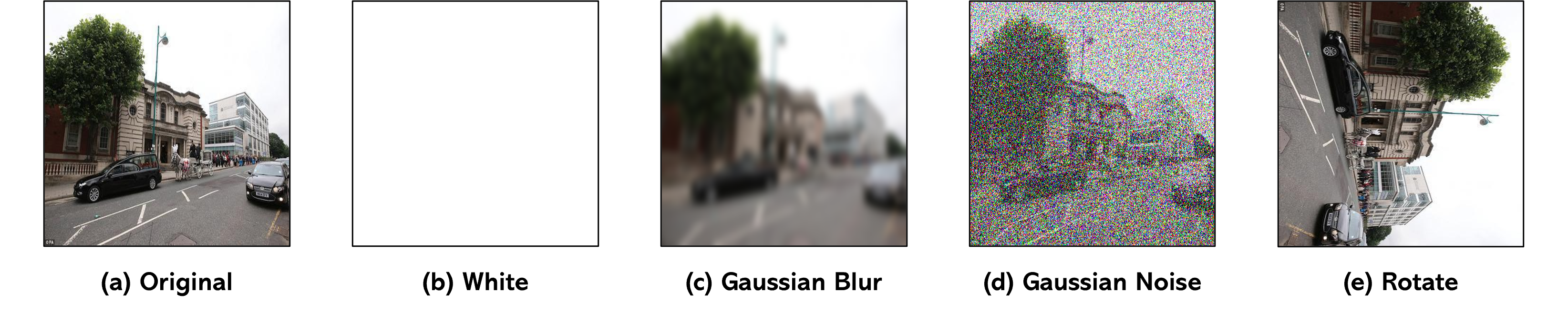}
    \caption{Visual Perturbation Strategies Evaluated for Skill Decomposition. To calculate the Visual Dependence Score ($V^{(i)}_k$, Eq. \ref{eq:vis_dependence}) , we compare the model's probability on the \textbf{(a) Original} image against four baselines: \textbf{(b) White} (Strategy adopted in main text), \textbf{(c) Gaussian Blur}, \textbf{(d) Gaussian Noise}, and \textbf{(e) Rotate}. Our analysis confirms that strategies which effectively destroy semantic information (b, c, d) yield high decomposition accuracy, whereas simple spatial transformation (e) preserves the visual content, leading to poor separation.}
    \label{fig:masking_ablation}
\end{figure}

In Eq. \ref{eq:vis_dependence} of the main text, we introduced the \textbf{Visual Dependence Score} ($V_k^{(i)}$) as the log-likelihood ratio between the model's probability given the original image $\mathbf{I}$ versus a non-informative baseline $\mathbf{I}_{\text{white}}$. In this section, we examine whether a blank white image the optimal baseline for measuring visual reliance.

Following the analysis in \citet{huang2025spotlight}, which examined \textit{token-level} visual dependency, we examine the optimal strategy to divide the visual grounding steps using our dynamic thresholding. Specifically, we evaluate four distinct perturbation methods to compute the reference probability $p_{w, k}^{(i)} = \log \pi_\theta(h_k^{(i)} | \mathbf{I}_{\text{perturbed}}, q, H_{<k})$. As illustrated in Figure~\ref{fig:masking_ablation}, these include:

\begin{enumerate}
    \item[(b)] \textbf{White Image (Strategy adopted in main text):} Replacing the visual input with a white image to remove all information.
    \item[(c)] \textbf{Gaussian Blur:} Applying a Gaussian blur kernel ($\sigma=5.0$) to remove high-frequency details while preserving color statistics.
    \item[(d)] \textbf{Gaussian Noise:} Adding random Gaussian noise ($\mu=0, \sigma=1$) to corrupt the signal structure.
    \item[(e)] \textbf{Rotate:} Rotating the input image by $90^\circ$ clockwise to alter spatial orientation.
\end{enumerate}

\paragraph{Results and Analysis.}
We evaluate each perturbation strategy on the \texttt{Qwen2.5-VL-7B-Instruct} backbone by measuring the skill decomposition accuracy achieved using our dynamic thresholding (Eq. \ref{eq:argmin_sse}). All results are validated against the ground-truth labeled dataset described in Appendix~\ref{app:gpt_label}. Our findings offer a clear insight into the nature of visual perception in large multimodal models:

\begin{itemize}
    \item \textbf{Information Destruction is Necessary:} The strategies that effectively destroy visual semantics (i.e., \textbf{White}, \textbf{Gaussian Blur}, and \textbf{Gaussian Noise}) perform similarly well, achieving decomposition accuracies in the range of \textbf{75.8\% - 76.2\%}. This suggests that the specific \textit{type} of noise matters less than the fact that the visual information is removed. We opted for the \textbf{White Image} in PDCR simply because it is the most computationally efficient and reproducible method to achieve this `blinding' effect.
    
    \item \textbf{Spatial Transformation is Insufficient:} In contrast, the \textbf{Rotation} strategy yields significantly poorer decomposition accuracy (\textbf{65.9\%}). This indicates that modern VLMs possess a degree of rotation invariance; even when the image is turned $90^\circ$, the model can still extract sufficient features to maintain a high likelihood for perception steps. Consequently, the log-likelihood ratio $V^{(i)}_k$ remains low for all steps, failing to distinguish perception from reasoning.
\end{itemize}

\noindent Thus, to accurately identify if a model is `seeing', we should distort the visual signal entirely rather than merely transforming it. This confirms that information-destroying baselines are essential for robust skill decomposition, validating our choice of the White Image strategy in our experiments.
\clearpage

\section{Qualitative Comparisons of Generated Reasoning}
\label{app: qual}
In this section, we present side-by-side comparisons between our proposed PDCR and baselines (GRPO \cite{grpo}, PACR \cite{yoon2025pacr}) on V-L reasoning tasks. All examples presented were generated by the final policies trained on the \texttt{Qwen2.5-VL-7B-Instruct} backbone. These examples highlight the core benefit of our decomposed reward signal: the mitigation of visual hallucinations. As observed in the figures below, baseline models often exhibit a specific failure mode where they generate logically coherent reasoning chains that are premised on incorrect visual extraction (highlighted in \textcolor{red}{\textbf{red}}). In contrast, PDCR guides the model to accurately ground its reasoning in the visual input (highlighted in \textcolor{teal}{\textbf{green}}). This accurate perception serves as a robust foundation for subsequent logic, leading to correct final answers in tasks involving spatial estimation, counting, and geometric reasoning.

\newtcolorbox{promptbox}[3][Judge Prompt]{%
  enhanced,
  breakable,
  colback=black!5!white,
  arc=5pt,
  boxrule=0.5pt,
  fonttitle=\bfseries,
  title=#1,
  before upper={\small},
  fontupper=\fontfamily{ptm}\selectfont,
  colframe=#2,                 
  colbacktitle=#2!70!black,    
  coltitle=white,             
  attach boxed title to top left={xshift=2mm, yshift*=-1mm},
  boxed title style={
    sharp corners,
    boxrule=0pt,
    arc=3pt,
    left=4pt,right=4pt,top=2pt,bottom=2pt
  },
  label=#3,
}
\definecolor{lightorange}{RGB}{253, 208, 162}
\definecolor{mintgreen}{RGB}{173, 223, 173}
\newcommand{\green}[1]{\textcolor{green}{#1}}

\begin{figure}[!p]
\centering
\vfill
\begin{promptbox}[Generated Sample 1 by GRPO]{lightorange}{prompt:orange}
\includegraphics[width=0.35\textwidth]{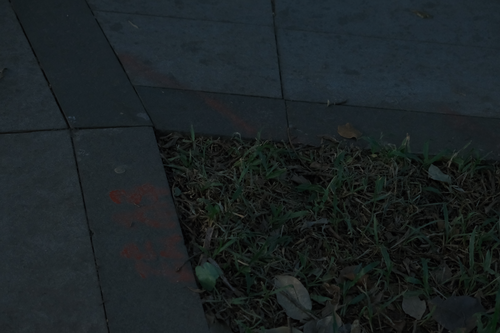}

\textbf{Question}: Which of the following options is a reasonable estimate of the angle between the ground surfaces in the figure?

\textbf{Choices}:
(a) 130 degrees \quad (b) 75 degrees \quad (c) 92 degrees \quad (d) 42 degrees

\colorbox{yellow!30}{\textbf{Ground truth: (b) 75 degrees}}

\textbf{Generation:}

\textbf{$<$think$>$} \\
To estimate the angle between the ground surfaces, we need to observe the intersection of the two surfaces. \red{The angle appears to be quite sharp, suggesting a smaller angle.} Let's analyze the options:

(a) 130 degrees: This is a very large angle, which is not typical for a ground surface intersection.

(b) 75 degrees: This is a moderate angle, but it's still relatively large.

(c) 92 degrees: This is a smaller angle, but it's still larger than what we typically see in ground surface intersections.

 (d) 42 degrees: This is a very small angle, which is more plausible for the intersection of two ground surfaces.

Given the visual observation, the angle appears to be closer to 42 degrees.
\textbf{$<$/think$>$}
\boxed{d}

\end{promptbox}
\vfill
\end{figure}

\begin{figure}[!p]
\centering
\vfill
\begin{promptbox}[Generated Sample 1 by PDCR (ours)]{mintgreen}{prompt:orange}
\includegraphics[width=0.35\textwidth]{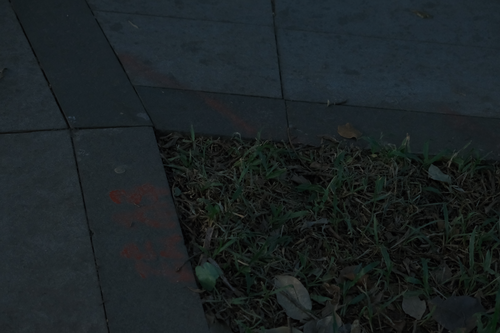}

\textbf{Question}: Which of the following options is a reasonable estimate of the angle between the ground surfaces in the figure?

\textbf{Choices}:
(a) 130 degrees \quad (b) 75 degrees \quad (c) 92 degrees \quad (d) 42 degrees

\colorbox{yellow!30}{\textbf{Ground truth: (b) 75 degrees}}

\textbf{Generation:}

\textbf{$<$think$>$} \\
To estimate the angle between the ground surfaces, we need to observe the intersection of the two surfaces. \pdcr{The angle appears to be less than 90 degrees but more than 45 degrees.} The angle looks closer to 75 degrees than to 130 degrees or 42 degrees. Therefore, the most reasonable estimate is 75 degrees.
\textbf{$<$/think$>$}
\boxed{b}

\end{promptbox}
\vfill
\end{figure}

\clearpage
\begin{figure}[!p]
\centering
\vfill
\begin{promptbox}[Generated Sample 2 by GRPO]{lightorange}{prompt:orange}
\includegraphics[width=0.20\textwidth]{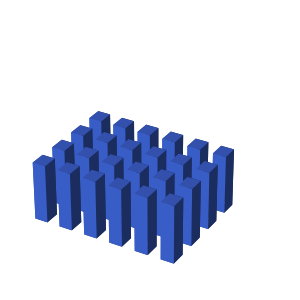}

\textbf{Question}: Which of the following options is a reasonable estimate of the number of cubes in the figure?

\textbf{Choices}:
(a) 17 \quad (b) 24 \quad (c) 7 \quad (d) 39

\colorbox{yellow!30}{\textbf{Ground truth: (b)}}

\textbf{Generation:}

\textbf{$<$think$>$} \\
To estimate the number of cubes in the figure, let's break it down step by step:

1. Visual Inspection: The figure appears to be a 3D structure with cubes stacked in a somewhat organized manner. It looks like a pyramid-like structure with a base and layers decreasing upwards.

2. Counting Layers:

    \red{
    - The base layer looks like it has 6 cubes.
    
    - The next layer up has 5 cubes.
    
    - The layer above that has 4 cubes.
    
    - The top layer has 3 cubes.
    }
    
3. Summing Up: Adding these together gives us 6 + 5 + 4 + 3 = 18 cubes.

4. Comparison with Options:

    (a) 17 is close but slightly less.
    
    (b) 24 is significantly more.
    
    (c) 7 is much less.
    
    (d) 39 is much more.

Given the visual inspection and the summing up, the closest reasonable estimate is 17.
\textbf{$<$/think$>$}
\boxed{a}

\end{promptbox}
\vfill
\end{figure}

\begin{figure}[!p]
\centering
\vfill
\begin{promptbox}[Generated Sample 2 by PDCR (ours)]{mintgreen}{prompt:orange}
\includegraphics[width=0.20\textwidth]{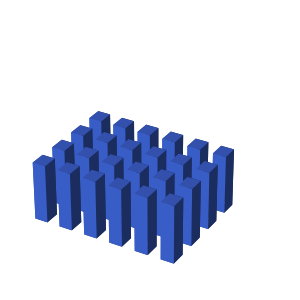}

\textbf{Question}: Which of the following options is a reasonable estimate of the number of cubes in the figure?

\textbf{Choices}:
(a) 17 \quad (b) 24 \quad (c) 7 \quad (d) 39

\colorbox{yellow!30}{\textbf{Ground truth: (b)}}

\textbf{Generation:}

\textbf{$<$think$>$} \\
To estimate the number of cubes in the figure, we can break it down into smaller, more manageable sections. The figure appears to be a 3D structure with a grid-like pattern. \pdcr{By counting the cubes in one row and one column, we can estimate the total number.}

\pdcr{- The figure has 6 rows.}

\pdcr{- Each row appears to have 4 cubes.}

So, the total number of cubes is approximately 6 rows * 4 cubes per row = 24 cubes.
\textbf{$<$/think$>$}
\boxed{b}

\end{promptbox}
\vfill
\end{figure}

\clearpage
\begin{figure}[!p]
\centering
\vfill
\begin{promptbox}[Generated Sample 3 by PACR]{lightorange}{prompt:orange}
\includegraphics[width=0.25\textwidth]{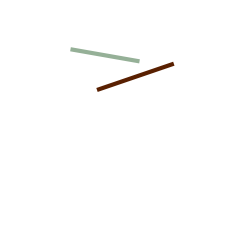}

\textbf{Question}: Which of the following options is a reasonable estimate of the sharp angle between two lines in the figure?

\textbf{Choices}:
(a) 67 degrees \quad (b) 51 degrees \quad (c) 29 degrees \quad (d) 85 degrees

\colorbox{yellow!30}{\textbf{Ground truth: (c) 29 degrees}}

\textbf{Generation:}

\textbf{$<$think$>$} \\
To estimate the sharp angle between the two lines, we need to visually assess the angle formed by the intersection of the two lines. \red{The angle appears to be less than 90 degrees but more than 45 degrees.}

- Option (a) 67 degrees is close to 90 degrees, which is not sharp.
- Option (b) 51 degrees is a reasonable estimate as it is less than 90 degrees and more than 45 degrees.
- Option (c) 29 degrees is too small for a sharp angle.
- Option (d) 85 degrees is close to 90 degrees, which is not sharp.

Therefore, the most reasonable estimate for the sharp angle between the two lines is 51 degrees.
\textbf{$<$/think$>$}
\boxed{b}

\end{promptbox}
\vfill
\end{figure}

\begin{figure}[!p]
\centering
\vfill
\begin{promptbox}[Generated Sample 3 by PDCR (ours)]{mintgreen}{prompt:orange}
\includegraphics[width=0.25\textwidth]{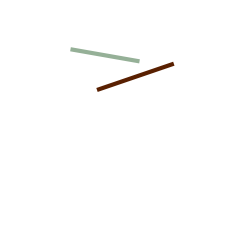}

\textbf{Question}: Which of the following options is a reasonable estimate of the sharp angle between two lines in the figure?

\textbf{Choices}:
(a) 67 degrees \quad (b) 51 degrees \quad (c) 29 degrees \quad (d) 85 degrees

\colorbox{yellow!30}{\textbf{Ground truth: (c) 29 degrees}}

\textbf{Generation:}

\textbf{$<$think$>$} \\
To estimate the sharp angle between the two lines, we need to consider the relative orientation of the lines. \pdcr{The lines appear to be intersecting at a point, and the angle between them is not a right angle (90 degrees) but rather a smaller angle less than 45 degrees.} Looking at the options provided, the angle that seems to be smaller than 45 degrees and thus the most reasonable for a sharp angle between two intersecting lines is 29 degrees.

Therefore, the most reasonable estimate of the sharp angle between the two lines is 29 degrees.
\textbf{$<$/think$>$}
\boxed{c}

\end{promptbox}
\vfill
\end{figure}

\clearpage

\begin{figure}[!p]
\centering
\vfill
\begin{promptbox}[Generated Sample 4 by PACR]{lightorange}{prompt:orange}
\includegraphics[width=0.25\textwidth]{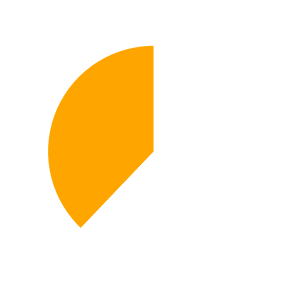}

\textbf{Question}: Which of the following options is a reasonable estimate of the area percentage of the orange portion of the pie chart?

\textbf{Choices}:
(a) (60\%, 70\%) \quad (b) (80\%, 90\%) \quad (c) (10\%, 20\%) \quad (d) (30\%, 40\%)

\colorbox{yellow!30}{\textbf{Ground truth: (d) (30\%, 40\%)}}

\textbf{Generation:}

\textbf{$<$think$>$} \\
The image shows a pie chart with a single orange section. \red{The orange section appears to cover less than a quarter of the entire circle, which is approximately 25\%. Given the options, the orange section is clearly less than 30\% of the pie chart.} Therefore, the most reasonable estimate for the area percentage of the orange portion is (c) (10\%, 20\%).
\textbf{$<$/think$>$}
\boxed{c}

\end{promptbox}
\vfill
\end{figure}

\begin{figure}[!p]
\centering
\vfill
\begin{promptbox}[Generated Sample 4 by PDCR (ours)]{mintgreen}{prompt:orange}
\includegraphics[width=0.25\textwidth]{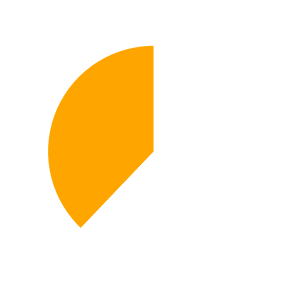}

\textbf{Question}: Which of the following options is a reasonable estimate of the area percentage of the orange portion of the pie chart?

\textbf{Choices}:
(a) (60\%, 70\%) \quad (b) (80\%, 90\%) \quad (c) (10\%, 20\%) \quad (d) (30\%, 40\%)

\colorbox{yellow!30}{\textbf{Ground truth: (d) (30\%, 40\%)}}

\textbf{Generation:}

\textbf{$<$think$>$} \\
\pdcr{The pie chart shows a large orange section that appears to cover more than a quarter but less than half of the circle.} A quarter of a circle is 25\%, and half is 50\%. The orange section looks to be between 30\% and 40\% of the circle. Therefore, the most reasonable estimate for the area percentage of the orange portion is (30\%, 40\%).
\textbf{$<$/think$>$}
\boxed{d}

\end{promptbox}
\vfill
\end{figure}

\clearpage

\clearpage

\section{Limitations and Future Works}
\label{sec:limitation}

Our work presents a promising direction for self-rewarding V-L models, but has limitations that offer avenues for future research.

\paragraph{Computational Cost.}
The primary limitation of our framework is the computational overhead during the training phase. As shown in Figure \ref{fig:train_curve}-(b) of main paper, both PDCR and the dense-reward baseline (PACR) \cite{yoon2025pacr} require additional computation to estimate the step-level confidence gains (and in our case, an additional visual dependence scores). This results in a higher per-step training cost compared to standard sparse-reward methods like GRPO \cite{grpo}. However, since our decomposition mechanism is lightweight, the cost of PDCR is comparable to that of PACR. 

Crucially, we argued that this training overhead is a favorable trade-off for inference efficiency. As shown in Figure \ref{fig:train_curve}-(c)\&(d), PDCR induces significantly more concise reasoning traces compared to GRPO. Consequently, while PDCR requires more GPU-hours to train, it yields a more efficient model for deployment. Future work could explore internal activation-based approximations to obtain these confidence scores without requiring full additional forward passes, potentially bridging the training cost gap.

\paragraph{Binary Skill Decomposition.}
Our method makes a simplifying assumption that V-L reasoning steps are a heterogeneous mixture of two \textit{distinct} and \textit{separable} skills. We then use Otsu's method to enforce a single, hard threshold $c^*$ between them. In reality, some steps may be hybrid (e.g., complex spatial reasoning that is both perceptual and logical). Our binary decomposition forces these hybrid steps into one of the two clusters, which may not be optimal. Future research could explore soft-clustering assignments or the identification of three or more distinct skill types.

%% file: pseudocode/pdcr.tex
\SetKwInOut{Input}{Input}
\SetKwInOut{Hyper}{Hyperparameters}
\SetKwProg{Fn}{Procedure}{}{}
\SetKwFunction{PDCRTrain}{PDCR\_Training\_Step}

\begin{algorithm}[h]
\caption{Perception-Decomposed Confidence Reward (PDCR) Training}
\label{alg:pdcr}
\small 
\SetKwInOut{Input}{Input}
\SetKwInOut{Hyper}{Hyperparameters}
\SetKwFunction{Otsu}{OtsuMethod}
\SetKwFunction{Sort}{Sort}
\SetKwFunction{Stats}{Stats}
\SetKwFunction{Normalize}{Normalize}


\Input{Current policy $\pi_\theta$, Reference policy $\pi_{\text{ref}}$,
       Training batch $\mathcal{D} = \{(\mathbf{I}_j, \mathbf{q}_j)\}_{j=1}^{B}$,
       \pdcr{White Image $\mathbf{I}_{\text{white}}$}}
\Hyper{Group size $N$, Discount $\gamma$, Weights $\lambda_O, \lambda_P$}

\BlankLine
\tcp{\textbf{Phase 1: Data Generation \& Decomposition}}
Initialize buffers: $\mathcal{T} \gets \emptyset$, $\mathcal{R} \gets \emptyset$\;
\pdcr{Initialize cluster pools: $\mathcal{I}_{\text{vis}} \gets \emptyset$, $\mathcal{I}_{\text{text}} \gets \emptyset$}\;

\ForEach{sample $(\mathbf{I}, \mathbf{q})$ in $\mathcal{D}$}{
    \For{$i \gets 1$ \KwTo $N$}{
        Generate trajectory $\tau^{(i)} = (h^{(i)}_1, \dots, h^{(i)}_{K_i},\hat{Y}^{(i)}) \sim \pi_{\theta}(\cdot \mid \mathbf{I}, \mathbf{q})$\;
        Compute outcome $R^{(i)} \gets is\_equivalent(\hat{Y}^{(i)},Y_{gt})$\;
        
        \For{$k \gets 1$ \KwTo $K_i$}{
            $g_k^{(i)} \gets \log \pi_\theta(Y_{gt} | \mathbf{I}, \mathbf{q}, H^{(i)}_{\leq k}) - \log \pi_\theta(Y_{gt}|\mathbf{I}, \mathbf{q}, H^{(i)}_{< k})$ \tcp*{Conf. Gain}
            
            \pdcr{$p_{\text{vis}} \gets \log \pi_\theta(h_k | \mathbf{I}, \mathbf{q}, H^{(i)}_{<k})$}\;
            \pdcr{$p_{\text{white}} \gets \log \pi_\theta(h_k | \mathbf{I}_{\text{white}}, \mathbf{q}, H^{(i)}_{<k})$}\;
            \pdcr{$V_k^{(i)} \gets p_{\text{vis}} - p_{\text{white}}$ \tcp*{Visual Dependence Score}}
        }
        
        \tcp{Per-Trajectory Skill Decomposition}
        \pdcr{$\mathbf{v}_{\text{sorted}} \gets \Sort(\{V_k^{(i)}\})$} \\
        \pdcr{$c^* \gets \Otsu(\mathbf{v}_{\text{sorted}})$ \tcp*{Dynamic Threshold (Eq. 8)}}
        
        \For{$k \gets 1$ \KwTo $K_i$}{
            \pdcr{\eIf{$V_k^{(i)} \ge c^*$}{
                Add $(i, k)$ to $\mathcal{I}_{\text{vis}}$\;
            }{
                Add $(i, k)$ to $\mathcal{I}_{\text{text}}$\;
            }}
        }
        Store $(\tau^{(i)}, \{g_k^{(i)}\})$ in $\mathcal{T}$, $R^{(i)}$ in $\mathcal{R}$\;
    }
}

\BlankLine
\tcp{\textbf{Phase 2: Decomposed Advantage Calculation}}
Compute sparse advantages $A_{\text{outcome}}^{(i)}$ using $\mathcal{R}$\;
Compute returns $\mathcal{G} = \{G_k^{(i)}\}$ from step gains $\{g_k^{(i)}\}$\;

\pdcr{Compute Stats: $(\text{min}_{\text{vis}}, \text{max}_{\text{vis}}) \gets \Stats(\{G_k^{(i)} \mid (i,k) \in \mathcal{I}_{\text{vis}}\})$}\;
\pdcr{Compute Stats: $(\text{min}_{\text{text}}, \text{max}_{\text{text}}) \gets \Stats(\{G_k^{(i)} \mid (i,k) \in \mathcal{I}_{\text{text}}\})$}\;

\ForEach{step $(i, k)$ in $\mathcal{T}$}{
    \pdcr{\eIf{$(i, k) \in \mathcal{I}_{\text{vis}}$}{
        $A_{proc,k}^{(i)} \gets \Normalize(G_k^{(i)}, \text{min}_{\text{vis}}, \text{max}_{\text{vis}})$\;
    }{
        $A_{proc,k}^{(i)} \gets \Normalize(G_k^{(i)}, \text{min}_{\text{text}}, \text{max}_{\text{text}})$\;
    }}
    $A_{total,k}^{(i)} \gets \lambda_O A_{\text{outcome}}^{(i)} + \lambda_P A_{proc,k}^{(i)}$\;
}

\BlankLine
\tcp{\textbf{Phase 3: Policy Update}}
$\mathcal{L}(\theta) \gets -\frac{1}{B \times N} \sum \sum A_{total,k}^{(i)} \min(\dots)$ \tcp*{GRPO Objective}
$\theta \gets \theta - \alpha \nabla_\theta \mathcal{L}(\theta)$\;

\end{algorithm}